\documentclass[conference]{IEEEtran}
\IEEEoverridecommandlockouts

\usepackage{cite}
\usepackage{amsmath,amssymb,amsfonts}
\usepackage{authblk}
\usepackage{graphicx}
\usepackage{booktabs} 
\usepackage[ruled,vlined,linesnumbered]{algorithm2e}
\usepackage{multirow}
\usepackage{balance}
\usepackage{subcaption}
\usepackage{balance}
\usepackage{footnote}
\usepackage{paralist}
\usepackage{enumitem}
\usepackage{algpseudocode}
\usepackage{xcolor, colortbl}

\definecolor{Gray}{gray}{0.75}
\newcolumntype{a}{>{\columncolor{Gray}}c}

\newcommand{\jian}[1]{{\small\color{brown}{\bf jian: #1}}}
\newcommand{\xintao}[1]{\textcolor{blue}{\emph{[Xintao: #1]}}}
\newcommand{\hide}[1]{}
\newcommand{\multifair}{\textsc{InfoFair}}

\newtheorem{prob}{Problem}
\newtheorem{lm}{Lemma}

\newtheorem{defn}{Definition}

\begin{document}
\title{\multifair: Information-Theoretic Intersectional Fairness}

\author[1]{Jian Kang}
\author[2]{Tiankai Xie}
\author[3]{Xintao Wu}
\author[2]{Ross Maciejewski}
\author[1]{Hanghang Tong}
\affil[1]{University of Illinois Urbana-Champaign, \{jiank2, htong\}@illinois.edu}
\affil[2]{Arizona State University, \{txie21, rmacieje\}@asu.edu}
\affil[3]{University of Arkansas, xintaowu@uark.edu}

\maketitle

\begin{abstract}
Algorithmic fairness is becoming increasingly important in data mining and machine learning. Among others, a foundational notation is {\em group fairness}. The vast majority of the existing works on group fairness, with a few exceptions, primarily focus on debiasing with respect to a single sensitive attribute, despite the fact that the co-existence of multiple sensitive attributes (e.g., gender, race, marital status, etc.) in the real-world is commonplace. As such, methods that can ensure a fair learning outcome with respect to all sensitive attributes of concern simultaneously need to be developed. In this paper, we study the problem of information-theoretic intersectional fairness
(\multifair), where statistical parity, a representative group fairness measure, is guaranteed among demographic groups formed by multiple sensitive attributes of interest. We formulate it as a mutual information minimization problem and propose a generic end-to-end algorithmic framework to solve it. The key idea is to leverage a variational representation of mutual information, which considers the variational distribution between learning outcomes and sensitive attributes, as well as the density ratio between the variational and the original distributions. Our proposed framework is generalizable to many different settings, including other statistical notions of fairness, and could handle any type of learning task equipped with a gradient-based optimizer. Empirical evaluations in the fair classification task on three real-world datasets demonstrate that our proposed framework can effectively debias the classification results with minimal impact to the classification accuracy.
\end{abstract}

\section{Introduction}\label{sec:intro}
The increasing amount of data and computational power have empowered machine learning algorithms to play crucial roles in automated decision-making for a variety of real-world applications, including 
credit scoring~\cite{luo2017deep}, criminal justice~\cite{berk2018fairness} and healthcare analysis~\cite{ahmad2018interpretable}. As the application landscape of machine learning continues to broaden and deepen, so does the concern regarding the potential, often unintentional, bias it could introduce or amplify. 
For example, recent media coverage has revealed that a well-trained image generator could turn a low-resolution picture of a black man into a high-resolution image of a white man due to the skewed data distribution that causes the model to disfavor the minority group,\footnote{https://www.theverge.com/21298762/face-depixelizer-ai-machine-learning-tool-pulse-stylegan-obama-bias} and another article highlighted an automated credit card application system assigning a dramatically higher credit limit to a man than to his female partner, even though his partner has a better credit history.\footnote{https://www.nytimes.com/2019/11/10/business/Apple-credit-card-investigation.html}

As such, algorithmic fairness, which aims to mitigate unintentional bias caused by automated learning algorithms, has become increasingly important. To date, researchers have proposed a variety of fairness notions~\cite{dwork2012fairness, feldman2015certifying}. Among them, one of the most fundamental notions is {\em group fairness}.\footnote{An orthogonal work in algorithmic fairness is individual fairness. Although it promises fairness by `treating similar individuals similarly' in principle, it is often hard to be operationalized in practice due to its strong assumption on distance metrics and data distributions.} Generally speaking, to ensure group fairness, the first step is to partition the entire population into a few demographic groups based on a pre-defined sensitive attribute (e.g., gender). Then the fair learning algorithm will enforce parity of a certain statistical measure among those demographic groups. Group fairness can be instantiated with many statistical notions of fairness. Statistical parity~\cite{zafar2017fairness} enforces the learned classifier to accept equal proportion of population from the pre-defined majority group and minority group. Likewise, disparate impact~\cite{feldman2015certifying} ensures the acceptance rate for the minority group should be no less than four-fifth of that for the group with the highest acceptance rate, which is analogous to the famous `four-fifth' rule in the legal support area~\cite{morris2000significance}. In addition, equalized odds and equal opportunity~\cite{hardt2016equality} are used to enforce the classification accuracies to be equal across all demographic groups conditioned on ground-truth outcomes or positively labeled populations, respectively. The vast majority of the existing works in group fairness primarily focus on debiasing with respect to a single sensitive attribute. However, it is quite common for multiple sensitive attributes (e.g., gender, race, marital status, etc.) to co-exist in a real-world application. 
We ask: {\em would a debiasing algorithm designed to ensure the group fairness for a particular sensitive attribute (e.g., marital status) unintentionally amplify the group bias with respect to another sensitive attribute (e.g., gender)? If so, how can we ensure a fair learning outcome with respect to all sensitive attributes of concern simultaneously?}

\begin{figure}
    \centering
    \includegraphics[width=.48\textwidth]{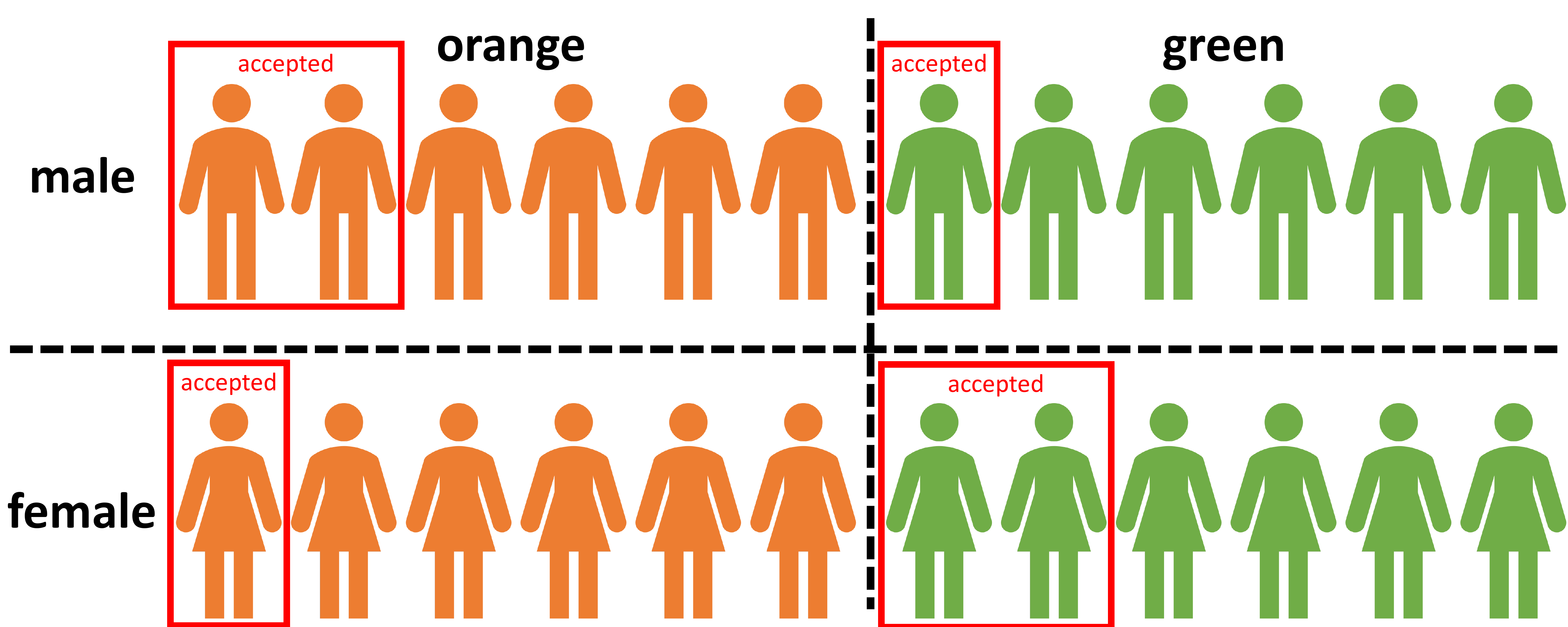}
    \vspace{-2mm}
    \caption[Caption for LOF]{An illustrative example of bias in job application classification when considering multiple sensitive attributes. Rows indicate gender (e.g., male vs. female) and columns indicate race (e.g., orange vs. green).\footnotemark Boxed individuals receive job offers. If we consider gender or race alone, statistical parity is enforced due to the equal acceptance rate. However, when considering gender and race (i.e., forming finer-grained gender-race groups), the classification result is biased in the fine-grained gender-race groups. This is because, the acceptance rates in two fine-grained groups (i.e., male-green group and female-orange) are lower than that of the two other fine-grained groups (i.e., male-orange and female-green).}
    \vspace{-6mm}
    \label{fig:example}
\end{figure}

Existing works for answering these questions~\cite{feldman2015certifying, zafar2017fairness, kearns2018preventing, bose2019compositional} have two major limitations. The first limitation is that some existing works could only debias multiple {\em distinct} sensitive attributes~\cite{bose2019compositional}, which fails to mitigate bias on the fine-grained groups formed by multiple sensitive attributes. Figure~\ref{fig:example} provides an illustrative example of the difference between fairness with respect to multiple distinct sensitive attributes and fairness among fine-grained groups of multiple sensitive attributes. The second limitation is that the optimization problems behind some existing works are often subject to surrogate constraints of statistical parity~\cite{feldman2015certifying, zafar2017fairness, kearns2018preventing} instead of directly optimizing statistical parity itself, resulting in unstable performance on bias mitigation unless the learned models could perfectly model the relationship between the training data and the ground-truth outcome.

In this paper, we tackle these two limitations by studying the problem of {\em information-theoretic intersectional fairness} 
(\multifair), which aims to directly enforce statistical parity on multiple sensitive attributes simultaneously. Though our focused fairness notion is statistical parity, the proposed method can be generalized to other statistical fairness notions (e.g., equalized odds and equal opportunity) with minor modifications. The key idea in solving the \multifair\ problem is to consider all sensitive attributes of interest as a vectorized sensitive attribute in order to partition the demographic groups and then minimize the dependence between learning outcomes and this vectorized attribute. More specifically, we measure the dependence using mutual information originated in information theory~\cite{shannon1948mathematical}. Building upon it, we formulate the \multifair\ problem as an optimization problem regularized on mutual information minimization. 

\footnotetext{We use imaginary race groups to avoid potential offenses.}

The main contributions of this paper are as follows.
\begin{compactitem}
	\item \textbf{Problem Definition.} We formally define the problem of information-theoretic intersectional fairness 
	and formulate it as an optimization problem, where the key idea is to minimize both the task-specific loss function (e.g., cross-entropy loss in classification) and mutual information between learning outcomes and the vectorized sensitive attribute.
	\item \textbf{End-to-End Algorithmic Framework.} We propose a novel end-to-end bias mitigation framework, named \multifair, by optimizing a variational representation of mutual information. The proposed framework is extensible and capable of solving any learning task with a gradient-based optimizer. 
	\item \textbf{Empirical Evaluations.} We perform empirical evaluations in the fair classification task on three real-world datasets. The evaluation results demonstrate that our proposed framework can effectively mitigate bias with little sacrifice in the classification accuracy. 
\end{compactitem}


\vspace{-2mm}
\section{Problem Definition}\label{sec:prob}
In this section, we present a table of the main symbols used in this paper. Then, we briefly review the concepts of statistical parity and mutual information, as well as their relationships. Finally, we formally define the problem of information-theoretic intersectional fairness.

\begin{table}[h!]
    \centering
    \vspace{-2mm}
    \caption{Table of symbols.}
    \vspace{-2mm}
    \begin{tabular}{|c|c|}
        \hline
        \textbf{Symbols} & \textbf{Definitions} \\
        \hline
        $\mathcal{D}$ & a set \\
        $\mathbf{W}$ & a matrix \\
        $\mathbf{h}$ & a vector \\
        $\mathbf{h}[i]$ & the $i$-th element in $\mathbf{h}$ \\
        \hline
        $\textrm{Pr}(\cdot)$ & the probability of an event happening \\
        $p_{\cdot, \cdot}$ & joint distribution of two random variables \\
        $p_{\cdot}$ & marginal distribution of a random variable \\
        $H(\cdot)$ & entropy \\
        $H(\cdot | \cdot)$ & conditional entropy \\
        $I(\cdot, \cdot)$ & mutual information \\
        \hline
    \end{tabular}
    \vspace{-2mm}
    \label{tab:symbols}
\end{table}

In this paper, matrices are denoted by bold uppercase letters (e.g., $\mathbf{X}$), vectors are denoted by bold lowercase letters (e.g., $\mathbf{y}$), scalars are denoted by italic lowercase letters (e.g., $c$) and sets are denoted by calligraphic letters (e.g., $\mathcal{D}$). We use superscript $^T$ to denote transpose (e.g., $\mathbf{h}^T$ is the transpose of $\mathbf{h}$) and superscript $\mathcal{C}$ to denote the complement of a set (e.g., set $\mathcal{D}^{\mathcal{C}}$ is the complement of set $\mathcal{D}$). We use a convention similar to NumPy for vector indexing (e.g., $\mathbf{h}[i]$ is the $i$-th element in vector $\mathbf{h}$).

\vspace{-3mm}
\subsection{Preliminaries}
\vspace{-1mm}
\noindent\textbf{Statistical Parity} is one of the most intuitive and widely-used group fairness notions. Given a set of data points $\mathcal{X}$, their corresponding labels $\mathbf{y}$ and a sensitive attribute $s$, classification with statistical parity aims to learn a classifier to predict outcomes that (1) are as accurate as possible with respect to $\mathbf{y}$ and (2) do not favor one group over another with respect to $s$. Mathematically,  statistical parity is defined as follows.

\begin{defn}\label{defn:statistical_parity}
(Statistical Parity~\cite{zafar2017fairness}). Suppose we have (1) a population $\mathcal{X}$, (2) a hypothesis $h:\mathcal{X}\rightarrow \{0,1\}$ which assigns a binary label to individual $x$ drawn from $\mathcal{X}$ and (3) a sensitive attribute which splits the population $\mathcal{X}$ into majority group $\mathcal{M}$ and minority group $\mathcal{M}^{\mathcal{C}}$ (i.e., $\mathcal{X} = \mathcal{M} \cup \mathcal{M}^{\mathcal{C}}$). An individual $x$ is accepted if $h(x)=1$ and rejected if $h(x)=0$. The hypothesis $h:\mathcal{X}\rightarrow \{0,1\}$ is said to have statistical parity on the population $\mathcal{X}$ as long as 
\vspace{-2mm}
\begin{equation}
\textrm{Pr}[h(x)=1|x\in\mathcal{M}] = \textrm{Pr}[h(x)=1|x\in\mathcal{M}^{\mathcal{C}}]
\vspace{-2mm}
\end{equation}
where $\textrm{Pr}[\cdot]$ denotes the probability of an event happening.
\end{defn}

Many methods have been proposed to achieve statistical parity. 
For example, Zemel et al.~\cite{zemel2013learning} learn fair representation by regularizing the difference in expected positive rate for majority and minority groups. 
Zhang et al.~\cite{zhang2018mitigating} propose an adversarial learning-based framework for fair classification, in which the output of the predictor is used to predict the sensitive attribute by the adversary. Kearns et al.~\cite{kearns2018preventing} propose a learner-auditor framework to enforce subgroup fairness through fictitious play strategy.

\noindent\textbf{Mutual Information} was first introduced in the 1940s~\cite{shannon1948mathematical}. Given two random variables, mutual information measures the dependence between them by quantifying the amount of information in bits obtained on one random variable through observing the other one. 
Let $(x,y)$ be a pair of random variables $x$ and $y$. Suppose their joint distribution is $p_{x,y}$ and the marginal distributions are $p_x$ and $p_y$. The mutual information between $x$ and $y$ is defined as
\vspace{-2mm}
\begin{equation}
\begin{aligned}
	I(x; y) & = H(x)-H(x|y) = \int_x \int_y p_{x,y}\log \frac{p_{x,y}}{p_x p_y} dx dy
\end{aligned}
\vspace{-1mm}
\end{equation}
where $H(x)=-\int_x p_x\log p_x dx$ is the entropy of $x$ and $H(x|y)=-\int_x\int_y p_{x,y} \log p_{x|y} dx dy$ is the conditional entropy of $x$ given $y$.
Unlike correlation coefficients (e.g., Pearson's correlation coefficient) which could only capture the linear dependence between two random variables, mutual information is more general in capturing both the linear and nonlinear dependence between two random variables. We have $I(x; y) = 0$ if and only if two random variables $x$ and $y$ are independent to each other.


According to Lemma~\ref{lm:sp_eq_mi}, there is an equivalence between statistical parity and zero mutual information.
\begin{lm}\label{lm:sp_eq_mi}
	(Equivalence between statistical parity and zero mutual information~\cite{zemel2013learning, ghassami2018fairness}). Statistical parity requires a sensitive attribute to be statistically independent to the learning results, which is equivalent to zero mutual information. Mathematically, given a learning outcome $\mathbf{\tilde y}$ and the sensitive attribute $s$, we have 
	\vspace{-2mm}
	\begin{equation}
		\underbrace{p_{\mathbf{\tilde y}|s} = p_{\mathbf{\tilde y}}}_\text{statistical parity}
		\Leftrightarrow
		p_{\mathbf{\tilde y}, s} = p_{\mathbf{\tilde y}}p_{s}
		\Leftrightarrow
		\underbrace{I(\mathbf{\tilde y}; s) = 0}_\text{zero mutual information}
	\end{equation}
\end{lm}
\vspace{-2mm}
\begin{IEEEproof}
	Omitted for brevity.
\end{IEEEproof}

\vspace{-2mm}
\subsection{Information-Theoretic Intersectional Fairness}
\vspace{-1mm}
In order to generalize Lemma~\ref{lm:sp_eq_mi} from a single sensitive attribute to a set of sensitive attributes $\mathcal{S} = \{s^{(1)}, \ldots, s^{(k)}\}$, we first introduce the concept of vectorized sensitive attribute $\mathbf{s}$ given $\mathcal{S}$. We define the vectorized sensitive attribute $\mathbf{s} = [s^{(1)}, \ldots, s^{(k)}]$ as a multi-dimensional random variable where each element of $\mathbf{s}$ represents the corresponding sensitive attribute in $\mathcal{S}$ (e.g., $\mathbf{s}[i] = s^{(i)}$ is the $i$-th sensitive attribute). Based on that, we have the following equivalence. For notational simplicity, we denote $I(\mathbf{\tilde y}; s^{(1)}, \ldots, s^{(k)})$, $p_{\mathbf{\tilde y}, s^{(1)}, \ldots, s^{(k)}}$ and $p_{s^{(1)}, \ldots, s^{(k)}}$ with $I(\mathbf{\tilde y}; \mathbf{s})$, $p_{\mathbf{\tilde y}, \mathbf{s}}$ and $p_{\mathbf{s}}$, respectively.

\vspace{-4mm}
\begin{equation}\label{eq:sp_eq_mi}
	p_{\mathbf{\tilde y}|\mathbf{s}} = p_{\mathbf{\tilde y}}
	\Leftrightarrow
	p_{\mathbf{\tilde y}, \mathbf{s}} = p_{\mathbf{\tilde y}}p_{\mathbf{s}}
	\Leftrightarrow
	I(\mathbf{\tilde y}; \mathbf{s}) = 0
	\vspace{-2mm}
\end{equation}

Based on Eq.~\eqref{eq:sp_eq_mi}, we formally define the problem of information-theoretic intersectional fairness as a mutual information minimization problem.

\begin{prob}\label{prob:multifair}
\multifair: Information-Theoretic Intersectional Fairness 
\end{prob}

\textbf{Input:} (1) a set of $k$ sensitive attributes $\mathcal{S} = \{s^{(1)}, \ldots,s^{(k)}\}$; (2) a set of $n$ data points $\mathcal{D}=\{(\mathbf{x}_i, \mathbf{s}_i, y_i)| i = 1,\ldots,n \}$ where $\mathbf{x}_i$ is the feature vector of the $i$-th data point, $y_i$ is its label and $\mathbf{s}_i = [s^{(1)}_i, \ldots, s^{(k)}_i]$ describes the vectorized sensitive attributes on $\mathcal{S}$ of the $i$-th data point (with $s^{(j)}_i$ being the corresponding attribute value of the $j$-th sensitive attribute $s^{(j)}$); and (3) a learning algorithm represented by $l(\mathbf{x}; \mathbf{s}; y; \mathbf{\tilde y}; \theta)$, where $l$ is the loss function, $\mathbf{\tilde y}^* = \textrm{argmin}_{\mathbf{\tilde y}} l(\mathbf{x}; \mathbf{s}; y; \mathbf{\tilde y}; \theta)$ is the optimal learning outcome on the input data with $\theta$ being model parameters.

\textbf{Output:} a set of revised learning outcomes $\{\mathbf{\tilde y}^{*}\}$ which minimizes (1) the empirical risk $\mathbb{E}_{(\mathbf{x}, \mathbf{s}, y)\sim \mathcal{D}}[l(\mathbf{x}; \mathbf{s}; y; \mathbf{\tilde y}; \theta)]$ and (2) the expectation of mutual information between the learning outcomes and the sensitive attributes $\mathbb{E}_{(\mathbf{x}, \mathbf{s}, y) \sim \mathcal{D}} \big[I(\mathbf{\tilde y}; \mathbf{s})\big]$.

\textbf{Remark:} a byproduct of \multifair\ is that the statistical parity can also be achieved on any subset of sensitive attributes included in $\mathcal{S}$, which is summarized in Lemma~\ref{lm:subset_fair}. This could be particularly useful in that the algorithm administrator does not need to re-train the model in order to obtain fair learning results if s/he is only interested in a subset of available sensitive attributes.
\begin{lm}\label{lm:subset_fair}
	Consider statistical parity as the fairness notion. Given a learning outcome $\mathbf{\tilde y}$, a set of $k$ sensitive attributes $\mathcal{S}=\{s^{(1)}, \ldots, s^{(k)}\}$ and the vectorized sensitive attribute $\mathbf{s} = [s^{(1)}, \ldots, s^{(k)}]$. If $\mathbf{\tilde y}$ is fair with respect to $\mathbf{s}$, then $\mathbf{\tilde y}$ is fair with respect to any vectorized sensitive attribute $\mathbf{s}_{\textrm{sub}}$ induced from the subset of sensitive attributes $\mathcal{S}_{\textrm{sub}} \subseteq \mathcal{S} = \{s^{(1)}, \ldots, s^{(k)}\}$.
\end{lm}

\begin{IEEEproof}
	Omitted for brevity. 
\end{IEEEproof}

\vspace{-2mm}
\section{Proposed Method}\label{sec:method}
\vspace{-1mm}
In this section, we present a generic end-to-end algorithmic framework, named \multifair, for 
information-theoretic intersectional fairness. 
We first formulate the problem as a mutual information minimization problem, 
and then present a variational representation of mutual information. Based on that, we present the \multifair\ framework to solve the optimization problem, followed by discussions on generalizations and variants of our proposed framework. 

\vspace{-2mm}
\subsection{Objective Function}
\vspace{-1mm}
Given a dataset $\mathcal{D}=\{(\mathbf{x}_i, \mathbf{s}_i, y_i) | i = 1, \ldots, n\}$, the \multifair\ problem (Problem~\ref{prob:multifair}) can be naturally formulated as minimizing the following objective function,
\vspace{-1.5mm}
\begin{equation}\label{eq:obj_mi}
    J = \mathbb{E}_{(\mathbf{x}, \mathbf{s}, y) \sim \mathcal{D}} \big[l(\mathbf{x}; \mathbf{s}; y; \mathbf{\tilde y}; \theta) + \alpha I(\mathbf{\tilde y}; \mathbf{s})\big]
    \vspace{-1.5mm}
\end{equation}
where $l$ is a task-specific loss function for a learning task, $\theta$ is the model parameter
, $\mathbf{\tilde y}$ is the learning outcome and $\alpha>0$ is the regularization hyperparameter. An example of loss function $l$ is the negative log likelihood
\vspace{-1.5mm}
\begin{equation}
l(\mathbf{x}; \mathbf{s}; y; \mathbf{\tilde y}; \theta) = -\log \mathbf{\tilde y}[y]
\vspace{-1.5mm}
\end{equation}
where $y$ is the class label and $\mathbf{\tilde y}$ denotes the probabilities of being classfied into the corresponding class.

To optimize the above objective function, a key challenge lies in optimizing the mutual information between the learning outcome and the vectorized sensitive feature $I(\mathbf{\tilde y}; \mathbf{s})$. Inspired by the seminal work of Belghazi et al.~\cite{belghazi2018mutual}, a natural choice would be to apply off-the-shelf mutual information estimation methods for high-dimensional data. Examples include MINE~\cite{belghazi2018mutual}, Deep Infomax~\cite{hjelm2018learning} and CCMI~\cite{mukherjee2020ccmi}, which estimate mutual information by parameterizing neural networks to maximize tight lower bounds of mutual information. However, in a mutual information minimization problem like Eq.~\eqref{eq:obj_mi}, it is often counter-intuitive to maximize a lower bound of mutual information. Though one could still maximize the objective function of these estimators to estimate the mutual information and use such estimation to guide the optimization of Eq.~\eqref{eq:obj_mi} as a minimax game, it is hindered by two hurdles. First, it requires learning a well-trained estimator to estimate the mutual information during each epoch of optimizing Eq.~\eqref{eq:obj_mi}. Second, if the estimator is not initialized with proper parameter settings, mutual information may be poorly estimated, which could further result in failing to find a good saddle point in such a minimax game.

\vspace{-2mm}
\subsection{Variational Representation of Mutual Information}\label{sec:variational_mi}
\vspace{-1mm}
In this paper, we take a different strategy from MINE and other similar methods by deriving a variational representation of mutual information $I(\mathbf{\tilde y}; \mathbf{s})$. Our variational representation leverages a variational distribution of the vectorized sensitive feature $\mathbf{s}$ given the learning outcome $\mathbf{\tilde y}$ (Lemma~\ref{lm:variational_mi}).

\begin{lm}\label{lm:variational_mi}
Suppose the joint distribution of the learning outcome $\mathbf{\tilde y}$ and the vectorized sensitive feature $\mathbf{s}$ is $p_{\mathbf{\tilde y}, \mathbf{s}}$ and the marginal distributions of $\mathbf{\tilde y}$ and $\mathbf{s}$ are $p_{\mathbf{\tilde y}}$ and $p_{\mathbf{s}}$, respectively. Mutual information $I(\mathbf{\tilde y}, \mathbf{s})$ between $\mathbf{\tilde y}$ and $\mathbf{s}$ is as follows.
\vspace{-2.5mm}
\begin{equation}
I(\mathbf{\tilde y}; \mathbf{s}) = H(\mathbf{s}) 
+ \mathbb{E}_{(\mathbf{\tilde y}, \mathbf{s})\sim p_{\mathbf{\tilde y}, \mathbf{s}}}
    \big[\log q_{\mathbf{s} | \mathbf{\tilde y}}\big] 
+ \mathbb{E}_{(\mathbf{\tilde y}, \mathbf{s})\sim p_{\mathbf{\tilde y}, \mathbf{s}}}
    \bigg[\log \frac{p_{\mathbf{\tilde y}, \mathbf{s}}}{p_{\mathbf{\tilde y}} q_{\mathbf{s} | \mathbf{\tilde y}}}\bigg]
\vspace{-2mm}
\end{equation}
where $q_{\mathbf{s}|\mathbf{\tilde y}}$ is the variational distribution of $\mathbf{s}$ given $\mathbf{\tilde y}$.
\end{lm}
\begin{IEEEproof}
    Omitted for brevity. 
\end{IEEEproof}

Next, we minimize the variational representation shown in Lemma~\ref{lm:variational_mi}, which contains three terms: 
(1) the entropy $H(\mathbf{s})$, (2) the expectation of log likelihood 
$\mathbb{E}_{(\mathbf{\tilde y}, \mathbf{s})\sim p_{\mathbf{\tilde y}, \mathbf{s}}} \big[\log q_{\mathbf{s} | \mathbf{\tilde y}}\big]$ and (3) the expectation of log density ratio $\mathbb{E}_{(\mathbf{\tilde y}, \mathbf{s})\sim p_{\mathbf{\tilde y}, \mathbf{s}}} \bigg[\log \frac{p_{\mathbf{\tilde y}, \mathbf{s}}}{p_{\mathbf{\tilde y}} q_{\mathbf{s} | \mathbf{\tilde y}}}\bigg]$. For the first term $H(\mathbf{s})$, we assume it to be a constant term, which can be ignored in the optimization stage. The rationale behind our assumption is that, in most (if not all) use cases, the vectorized sensitive feature $\mathbf{s}$ relates to the demographic information of an individual (e.g., gender, race, marital status, etc.), which should remain unchanged during the learning process. Then the remaining key challenges lie in (C1) calculating $\log q_{\mathbf{s}|\mathbf{\tilde y}}$ and (C2) estimating $\log \frac{p_{\mathbf{\tilde y}, \mathbf{s}}}{p_{\mathbf{\tilde y}} q_{\mathbf{s}|\mathbf{\tilde y}}}$. The intuition of C1 and C2 is that we strive to find a learning outcome $\mathbf{\tilde y}$ such that (1) $\mathbf{\tilde y}$ fails to predict the vectorized sensitive feature $\mathbf{s}$ (refer to C1), while (2) making it hard to distinguish if the vectorized sensitive feature $\mathbf{s}$ is generated from the variational distribution or sampled from the original distribution (refer to C2).

\noindent \textbf{C1 -- Calculating $\log q_{\mathbf{s}|\mathbf{\tilde y}}$}. It can be naturally formulated as a prediction problem, where the input is the learning outcome $\mathbf{\tilde y}$ and the output is the probability of $\mathbf{s}$ being predicted. To solve it, we parameterize a decoder $f(\mathbf{\tilde y}; \mathbf{s}; \mathbf{W})$ (e.g., a neural network) as a sensitive feature predictor to `reconstruct' $\mathbf{s}$, where $\mathbf{W}$ is the learnable parameters in the decoder.
\vspace{-2mm}
\begin{equation}\label{eq:sensitivie_decoder}
    \log q_{\mathbf{s}|\mathbf{\tilde y}} = \log f(\mathbf{\tilde y}; \mathbf{s}; \mathbf{W})
    \vspace{-2mm}
\end{equation}

For categorical sensitive attribute, $\log q_{\mathbf{s}|\mathbf{\tilde y}}$ refers to the log likelihood of classifying $\mathbf{\tilde y}$ into label $\mathbf{s}$, which can be interpreted as the negative of cross-entropy loss of the decoder $f(\mathbf{\tilde y}; \mathbf{s}; \mathbf{W})$. Moreover, if $\mathbf{s}$ contains multiple categorical sensitive attributes, solving Eq.~\eqref{eq:sensitivie_decoder} requires solving a multi-label classification problem, which itself is not trivial to solve. In this case, we further reduce it to a single-label problem by applying a mapping function $\textit{map}()$ to map the multi-hot encoding $\mathbf{s}$ into a one-hot encoding $\mathbf{\hat s}$ (i.e., $\mathbf{\hat s} = \textit{map}(\mathbf{s})$).

\noindent \textbf{C2 -- Estimating $\log \frac{p_{\mathbf{\tilde y}, \mathbf{s}}}{p_{\mathbf{\tilde y}} q_{\mathbf{s}|\mathbf{\tilde y}}}$}. In practice, calculating $p_{\mathbf{\tilde y}, \mathbf{s}}$ and $p_{\mathbf{\tilde y}} q_{\mathbf{s}|\mathbf{\tilde y}}$ individually is hard since the underlying distributions $p_{\mathbf{\tilde y}, \mathbf{s}}$ and $p_{\mathbf{\tilde y}}$ are often unknown. Recall that our goal is to estimate the log of the ratio between these two joint distributions. Therefore, we estimate it through {\em density ratio estimation}, where the numerator $p_{\mathbf{\tilde y}, \mathbf{s}}$ denotes the original joint distribution of the learning outcome $\mathbf{\tilde y}$ and ground-truth vectorized sensitive feature $\mathbf{s}$, and the denominator $p_{\mathbf{\tilde y}} q_{\mathbf{s}|\mathbf{\tilde y}}$ denotes the joint distribution of the learning outcome $\mathbf{\tilde y}$ and the vectorized sensitive feature $\mathbf{\tilde s}$ generated from the learning outcome using the aforementioned decoder. 

We further reduce this density ratio estimation problem to a class probability estimation problem, which was originally developed in  \cite{bickel2009discriminative} for solving a different problem (i.e., the classification problem with the input distribution and the test distribution differing arbitrarily). The core idea is that, given a pair of learning outcome and vectorized sensitive feature, we want to predict whether it is drawn from the original joint distribution or from the joint distribution inferred by the decoder. We label each pair of learning outcome and ground-truth vectorized sensitive feature $(\mathbf{\tilde y}, \mathbf{s})$ with a positive label ($c=1$) and each pair of learning outcome and generated vectorized sensitive feature $(\mathbf{\tilde y}, \mathbf{\tilde s})$ with a negative label ($c=-1$). After that, we rewrite the probability densities as $p_{\mathbf{\tilde y}, \mathbf{s}} = \textrm{Pr}[c = 1 | \mathbf{\tilde y}, \mathbf{s}]$ and $p_{\mathbf{\tilde y}} q_{\mathbf{s}|\mathbf{\tilde y}} = \textrm{Pr}[c = -1 | \mathbf{\tilde y}, \mathbf{s}]$.
Then the density ratio can be further rewritten as 
\vspace{-1.5mm}
\begin{equation}\label{eq:dre_to_cpe}
\begin{aligned}
    \log\frac{p_{\mathbf{\tilde y}, \mathbf{s}}}{p_{\mathbf{\tilde y}} q_{\mathbf{s}|\mathbf{\tilde y}}} & = \log \frac{\textrm{Pr}[c = 1 | \mathbf{\tilde y}, \mathbf{s}]}{\textrm{Pr}[c = -1 | \mathbf{\tilde y}, \mathbf{s}]} = \textit{logit}(\textrm{Pr}[c = 1 | \mathbf{\tilde y}, \mathbf{s}])
\end{aligned}
\vspace{-1.5mm}
\end{equation}
Furthermore, if we model $\textrm{Pr}[c = 1 | \mathbf{\tilde y}, \mathbf{s}]$ using logistic regression (i.e., $\textrm{Pr}[c = 1 | \mathbf{\tilde y}, \mathbf{s}] = \textit{logistic}(\mathbf{\tilde y}, \mathbf{s})$), Eq.~\eqref{eq:dre_to_cpe} is reduced to a simple linear function as
\vspace{-1.5mm}
\begin{equation}\label{eq:dre_to_linear}
    \log\frac{p_{\mathbf{\tilde y}, \mathbf{s}}}{p_{\mathbf{\tilde y}} q_{\mathbf{s}|\mathbf{\tilde y}}} = \textit{logit}(\textit{logistic}(\mathbf{\tilde y}, \mathbf{s})) = \mathbf{w}_1^T\mathbf{\tilde y} + \mathbf{w}_2^T\mathbf{s}
    \vspace{-1.5mm}
\end{equation}
where both $\mathbf{w}_1$ and $\mathbf{w}_2$ are learnable parameters. Putting everything together, we rewrite Eq.~\eqref{eq:obj_mi} as 
\vspace{-1.5mm}
\begin{equation}\label{eq:obj_variational}
\begin{aligned}
    J =\ & \mathbb{E}_{(\mathbf{x}, y) \sim \mathcal{D}} 
            \big[l(\mathbf{x}; \mathbf{s}; y; \mathbf{\tilde y}; \theta) 
            + \alpha \log q_{\mathbf{s}|\mathbf{\tilde y}}\big] \\
        & + \alpha \mathbb{E}_{\{(\mathbf{\tilde y}, \mathbf{s}) \sim p_{\mathbf{\tilde y}, \mathbf{s}}\} \cup \{(\mathbf{\tilde y}, \mathbf{s}) \sim p_{\mathbf{\tilde y}} q_{\mathbf{s}|\mathbf{\tilde y}}\}} 
            \big[\mathbf{w}_1^T\mathbf{\tilde y} + \mathbf{w}_2^T\mathbf{s}\big]
\end{aligned}
\vspace{-1.5mm}
\end{equation}
where $p_{\mathbf{\tilde y}, \mathbf{s}}$ is the joint distribution of the learning outcome $\mathbf{\tilde y}$ and ground-truth vectorized sensitive feature $\mathbf{s}$, $p_{\mathbf{\tilde y}} q_{\mathbf{s}|\mathbf{\tilde y}}$ is the joint distribution of the learning outcome $\mathbf{\tilde y}$ and predicted vectorized sensitive feature $\mathbf{s}$.

\begin{figure*}
    \centering
    \includegraphics[width=.7\textwidth]{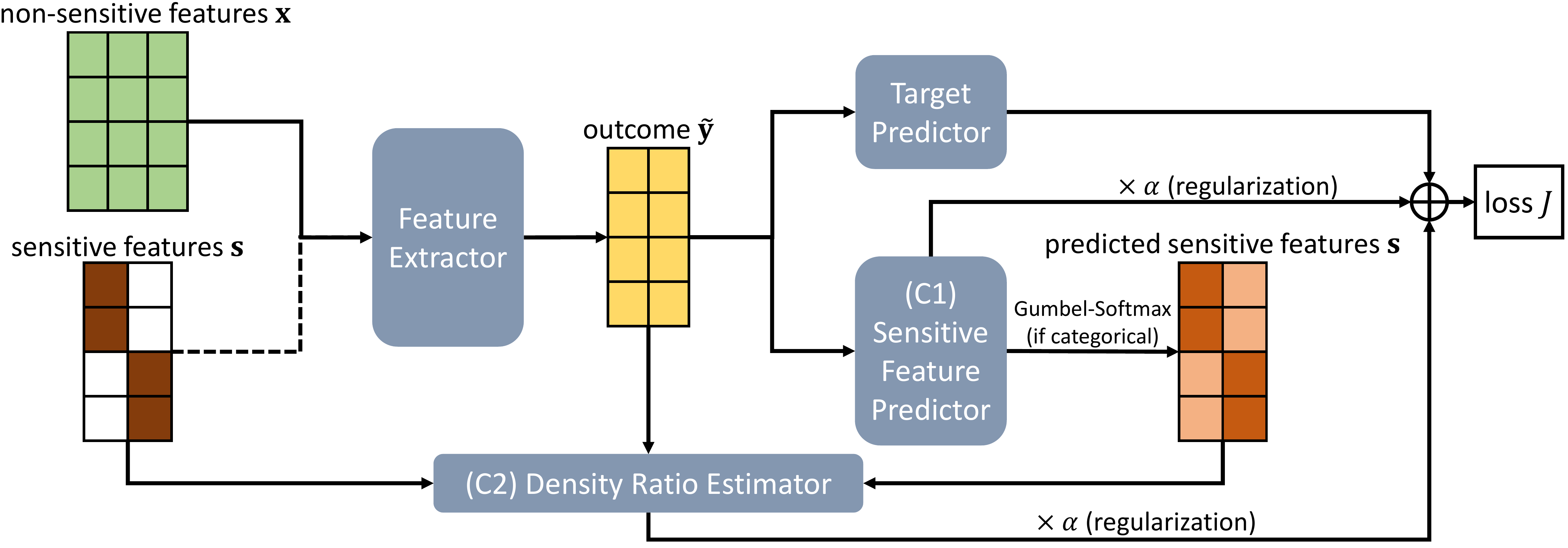}
    \caption{A General overview of our proposed \multifair\ framework. The dashed line between sensitive feature $\mathbf{s}$ and feature extractor means that sensitive features can be optionally passed into feature extractor as the input.}
    \vspace{-6mm}
    \label{fig:model}
\end{figure*}

\vspace{-2mm}
\subsection{\multifair: Overall Framework}\label{sec:framework}
\vspace{-1mm}
Based on the objective function (Eq.~\eqref{eq:obj_variational}), we propose a generic end-to-end framework to solve the information-theoretic intersectional fairness 
problem. A general overview of the model architecture is shown in Fig.~\ref{fig:model}. Our proposed model contains four main modules, including (1) feature extractor, (2) target predictor, (3) sensitive feature predictor and (4) a density ratio estimator. 
In principle, as long as each module is differentiable, the proposed framework can be optimized by any gradient-based optimizer. 

The general workflow of \multifair\ is as follows. 
\begin{compactitem}
    \item [1.] The non-sensitive features and sensitive features (optional) are passed into a feature extractor to extract the learning outcomes;
    \item [2.] The learning outcomes will be fed into a target predictor to predict the targets for a certain downstream task (i.e., $l(\mathbf{x}; \mathbf{s}; y; \mathbf{\tilde y}; \theta)$ in Eq.~\eqref{eq:obj_variational});
    \item [3.] The learning outcomes will be passed into the sensitive feature predictor to `reconstruct' the vectorized sensitive features (i.e., $\log q_{\mathbf{s}|\mathbf{\tilde y}}$ in Eq.~\eqref{eq:obj_variational});
    \item [4.] Together with the learning outcomes and the ground-truth vectorized sensitive features, the predicted vectorized sensitive features will be used to estimate the density ratio between the original distribution and the variational distribution (i.e., $\mathbf{w}_1^T\mathbf{\tilde y} + \mathbf{w}_2^T\mathbf{s}$ in Eq.~\eqref{eq:obj_variational}).
\end{compactitem}

Given a data point with categorical sensitive attribute(s), the predicted vectorized sensitive feature $\mathbf{s}$ is usually denoted as a one-hot vector. However, learning a one-hot vector is a difficult problem due to the discrete nature of vector elements, which makes the computation non-differentiable. To address this issue, we approximate such one-hot encoding by Gumbel-Softmax~\cite{jang2017categorical}, which can be calculated as $\mathbf{s}[i] = \frac{\exp(\big[\log(\mathbf{o}_{\mathbf{s}}[i]) + g_i\big] / \tau)}{\sum_{j=1}^{n_\mathbf{s}}\exp(\big[\log(\mathbf{o}_{\mathbf{s}}[j]) + g_j\big] / \tau)}$, where $\mathbf{o}_{\mathbf{s}}$ is the output of the sensitive feature predictor, $n_\mathbf{s}$ is the dimension of $\mathbf{s}$, $g_1,\ldots,g_{n_\mathbf{s}}$ are i.i.d. points drawn from $\textrm{Gumbel}(0, 1)$ distribution, and $\tau$ is the softmax temperature. As $\tau \rightarrow \infty$, the Gumbel-Softmax samples are uniformly distributed; while as $\tau \rightarrow 0$, the Gumbel-Softmax distribution converges to a one-hot categorical distribution. In \multifair, we start with a high temperature and then anneal it during epochs of training.

\vspace{-2mm}
\subsection{\multifair: Generalizations and Variants}\label{sec:generalization}
\vspace{-1mm}
The proposed \multifair\ is able to be generalized in multiple aspects. Due to the space limitation, we only give some brief descriptions, each of which could be a future direction in applying our proposed framework.

\noindent \textbf{A -- \multifair\ with equal opportunity.} Our \multifair\ framework is generalizable to enforce equal opportunity~\cite{hardt2016equality}, another widely-used group fairness notions. We leave for future work to explore the potential of \multifair\ in enforcing equal opportunity.

Equal opportunity ensures equality across demographic groups for a preferred label (i.e., the label that benefits an individual). Mathematically, it is defined as follows.

\begin{defn}\label{defn:equal_opportunity}
(Equal Opportunity~\cite{hardt2016equality}). 
Following the settings of Definition~\ref{defn:statistical_parity}, if equal opportunity is enforced, the hypothesis $h:\mathcal{X}\rightarrow \{0,1\}$ satisfies
\vspace{-2mm}
\begin{equation}
\textrm{Pr}[h(x)=1|x\in\mathcal{M}, y=1] = \textrm{Pr}[h(x)=1|x\in\mathcal{M}^{\mathcal{C}}, y=1]
\end{equation}
where $\textrm{Pr}[\cdot]$ denotes the probability of an event happening.
\end{defn}

Analogous to the relationship between mutual information and statistical parity, ensuring equal opportunity is essentially a conditional mutual information minimization problem.
\vspace{-2mm}
\begin{equation}
    \underbrace{p_{\mathbf{\tilde y}|\mathbf{s}, y=1} = p_{\mathbf{\tilde y}|y=1}}_\text{equal opportunity}
	\Leftrightarrow
	\underbrace{I(\mathbf{\tilde y}; \mathbf{s}|y=1) = 0}_\text{zero conditional mutual information}
	\vspace{-2mm}
\end{equation}
By the definition of conditional mutual information, we have $I(\mathbf{\tilde y}; \mathbf{s}|y=1) = H(\mathbf{s}|y=1) - H(\mathbf{s}|\mathbf{\tilde y}, y=1)$. For $H(\mathbf{s}|y=1)$, we assume it as a constant term by the similar rationale of statistical parity. 
Similarly, we can rewrite $H(\mathbf{s}|\mathbf{\tilde y}, y=1)$ as
\vspace{-2mm}
\begin{equation}
\begin{aligned}
H(\mathbf{s}|\mathbf{\tilde y}, y=1) = 
& \mathbb{E}_{(\mathbf{\tilde y}, \mathbf{s})\sim p_{\mathbf{\tilde y}, \mathbf{s}|y=1}} \big[
    -\log q_{\mathbf{s}|\mathbf{\tilde y}, y=1}
\big] \\
& - \mathbb{E}_{(\mathbf{\tilde y}, \mathbf{s})\sim p_{\mathbf{\tilde y}, \mathbf{s}|y=1}} \bigg[
    \log \frac{p_{\mathbf{\tilde y}, \mathbf{s}|y=1}}{p_{\mathbf{\tilde y}|y=1} q_{\mathbf{s}|\mathbf{\tilde y},y=1}}
\bigg]
\end{aligned}
\vspace{-1mm}
\end{equation}
Then, to compute $\log q_{\mathbf{s}|\mathbf{\tilde y}, y=1}$, we could adopt similar strategy as computing $\log q_{\mathbf{s}|\mathbf{\tilde y}}$ in Section~\ref{sec:variational_mi} by constructing a decoder $f(\mathbf{\tilde y}, \mathbf{s}, \mathbf{W})$ to `reconstruct' $\mathbf{s}$ for {\em positive training samples}. Similarly, estimating the density ratio 
can be achieved by applying Eq.~\eqref{eq:dre_to_linear} on {\em positive training samples}. Thus, \multifair\ is able to enforce equal opportunity by minimizing
\begin{equation}\label{eq:obj_variational_eo}
\begin{aligned}
    J =\ & \mathbb{E}_{(\mathbf{x}, y) \sim \mathcal{D}} 
            \big[l(\mathbf{x}; \mathbf{s}; y; \mathbf{\tilde y}; \theta) 
            + \alpha \log q_{\mathbf{s}|\mathbf{\tilde y}}\big] \\
        & + \alpha \mathbb{E}_{\{(\mathbf{\tilde y}, \mathbf{s}) \sim p_{\mathbf{\tilde y}, \mathbf{s}|y=1}\} \cup \{(\mathbf{\tilde y}, \mathbf{s}) \sim p_{\mathbf{\tilde y}|y=1} q_{\mathbf{s}|\mathbf{\tilde y}, y=1}\}} 
            \big[\mathbf{w}_1^T\mathbf{\tilde y} + \mathbf{w}_2^T\mathbf{s}\big]
\end{aligned}
\end{equation}

\noindent \textbf{B -- Relationship to adversarial debiasing.} Adversarial debiasing framework~\cite{zhang2018mitigating} consists of (1) a predictor that predicts the class membership probabilities using given data and (2) an adversary that takes the output of the predictor to predict the sensitive attribute of given data. The framework is optimized to minimize the loss function of the predictor while maximizing the loss function of the adversary. If we merge feature extractor and target predictor to one single module and remove the density ratio estimator, \multifair\ will degenerate to the adversarial debiasing method.

\noindent \textbf{C -- Relationship to Information Bottleneck.} If we set the loss function $l$ in Eq.~\eqref{eq:obj_mi} as the negative mutual information $-I(\mathbf{\tilde y}; y)$, Eq.~\eqref{eq:obj_mi} becomes the information bottleneck method~\cite{tishby2000information}. Then the goal becomes to learn $\mathbf{\tilde y}$ that depends on the vectorized sensitive attribute $\mathbf{s}$ minimally and ground truth $y$ maximally.

\noindent \textbf{D -- Fairness for continuous-valued sensitive features.} 
Most existing works in fair machine learning only consider categorical sensitive attribute (e.g., gender, race). Our proposed \multifair\ framework could be generalized to continuous-valued features as mutual information supports continuous-valued random variables. This advantage could empower our framework to work in even more application scenarios. For example, in image classification, we can classify images without the impact of certain image patches (e.g., patches that relate to individual's skin color). However, a major difficulty lies in modeling the variational distribution of sensitive attribute given the learning outcomes extracted from feature extractor. A potential solution could be utilizing a generative model (e.g., VAEs~\cite{kingma2013auto}) as the sensitive feature predictor.

\noindent \textbf{E -- Fairness for non-i.i.d. graph data.} For fair graph mining, given a graph $G = (\mathbf{A}, \mathbf{X})$ where $\mathbf{A}$ is the adjacency matrix and $\mathbf{X}$ is the node feature matrix, we can use graph convolutional layer(s) as a feature extractor with the weight of the last layer to be identity matrix $\mathbf{I}$ and no nonlinear activation in the last graph convolution layer, in order to extract node representations. The reason for such a specific architecture in the last graph convolution layer is as follows. In general, a graph convolutional layer consists of two operations: feature aggregation $\mathbf{Z} = f_{\textrm{aggregate}}(\mathbf{A}; \mathbf{X}) = \mathbf{A} \mathbf{X}$ and feature transformation $\mathbf{H} = f_{\textrm{transform}}(\mathbf{Z}; \mathbf{W}) = \sigma(\mathbf{Z}\mathbf{W})$ where $\mathbf{W}$ is learnable parameters and $\sigma$ is usually a nonlinear activation. The last layer in GCN~\cite{kipf2017semi} is simply $\textit{softmax}(\mathbf{A}\mathbf{X}\mathbf{W})$, which can be viewed as a general multi-class logistic regression on the aggregated feature $\mathbf{Z} = \mathbf{A}\mathbf{X}$ (i.e., $\textit{softmax}(\mathbf{Z}\mathbf{W})$). 

\noindent \textbf{F -- Fairness beyond classification.} Note that \multifair\ does not have specific restrictions on the architecture of the feature extractor, target predictor or sensitive target predictor, which empowers it to handle many different types of tasks by selecting the proper architecture for each module. For example, if an analyst aims to learn fair representations with respect to gender for recommendation, s/he can set the feature extractor to be a multi-layer perceptron (MLP) for learning outcome extraction, the target predictor layer to be an MLP that predicts a rating and minimizes the mean squared error (MSE) between the predicted rating and ground-truth rating, and the sensitive target predictor to be another MLP with softmax to predict the gender based on extracted embedding.

\vspace{-2mm}
\section{Experimental Evaluation}\label{sec:exp}
\vspace{-1mm}
In this section, we conduct experimental evaluations. All experiments are designed to answer the following questions:
\begin{compactitem}
    \item [\textbf{RQ1.}] How does the fairness impact the learning performance?
    \item [\textbf{RQ2.}] How effective is \multifair\ in mitigating bias?
\end{compactitem}

\vspace{-3mm}
\subsection{Experimental Settings}
\vspace{-1mm}
\noindent{\bf A -- Datasets.} We test the proposed method on three commonly-used datasets in fair machine learning research. The statistics of these datasets are summarized in Table~\ref{tab:datasets}.

\begin{table}[h!]
    \centering
    \vspace{-2mm}
    \caption{Statistics of datasets.}
    \vspace{-2mm}
    \begin{tabular}{|c|c|c|c|}
        \hline
        \textbf{Datasets} & \textbf{\# Samples} & \textbf{\# Attributes} & \textbf{\# Classes} \\
        \hline
        COMPAS & 6,172 & 52 & 2 \\ 
        Adult Income & 45,222 & 14 & 2 \\  
        Dutch Census & 60,420 & 11 & 2 \\  
        \hline
    \end{tabular}
    \vspace{-4mm}
    \label{tab:datasets}
\end{table}

\hide{For all datasets, we randomly split them into 80\% training set and 20\% test set. A description of each dataset is shown below. We provide additional details on data preprocessing in the Appendix. 

\begin{itemize}
	 \item \textit{COMPAS} dataset contains in total of 6,172 criminal defendants in Broward County, Florida. Each defendant is described by 52 attributes used by the COMPAS (Correctional Offender Management Profiling for Alternative Sanctions) algorithm for scoring their likelihood of reoffending crimes in the following 2 years. The goal is to determine whether a criminal defendant will reoffend in the next 2 years.
	 \item \textit{Adult Income} dataset contains in total of 45,222 individuals. Each individual is described by 14 attributes that relate to his/her personal demographic information, including gender, race, education, marital status, etc. The goal is to predict whether a person can earn a salary over \$50,000 a year. 
    \item \textit{Dutch Census} dataset contains in total of 60,420 individuals.\footnote{https://sites.google.com/site/faisalkamiran/} Each individual is described by 11 attributes that relate to his/her demographic and economic information to predict whether s/he has a prestigious occupation.
\end{itemize}}

\noindent{\bf B -- Baseline Methods.} We compare \multifair\ with several baseline methods, including \textit{Learning Fair Representations (LFR)}~\cite{zemel2013learning}, \textit{MinDiff}~\cite{prost2019toward}, \textit{Generalized Demographic Parity (GDP)}~\cite{jiang2021generalized}, \textit{Adversarial Debiasing (Adversarial)}~\cite{zhang2018mitigating}, \textit{Fair Classification with Fairness Constraints (FCFC)}~\cite{zafar2017fairness}, \textit{GerryFair}~\cite{kearns2018preventing} and \textit{Disparate Impact (DI)}~\cite{feldman2015certifying}. 

\setlength{\aboverulesep}{0pt}
\setlength{\belowrulesep}{0pt}
\begin{table*}
	\centering
	\caption{Debiasing results on all datasets. Lower is better for the gray column (Imparity). Higher is better for all others.}
	\vspace{-2mm}
	\label{tab:effectiveness}
	\resizebox{.98\linewidth}{!}{
	\begin{tabular}{|c|c|a|c|c|a|c|c|a|c|}
	    \hline
		\multicolumn{10}{|c|}{\textbf{Debiasing results on {\em COMPAS} dataset}} \\
		\hline
		\multirow{2}{*}{\textbf{Method}} & \multicolumn{3}{c|}{\textbf{gender}} & \multicolumn{3}{c|}{\textbf{race}} & \multicolumn{3}{c|}{\textbf{gender \& race}} \\
		\cmidrule{2-10}
		& \textbf{Micro/Macro F1} & \textbf{Imparity} & \textbf{Reduction}
		& \textbf{Micro/Macro F1} & \textbf{Imparity} & \textbf{Reduction}
		& \textbf{Micro/Macro F1} & \textbf{Imparity} & \textbf{Reduction} \\
		\hline
		\textbf{Vanilla} & $0.972/0.972$ & $0.050$ & $0.000\%$
		                 & $0.972/0.972$ & $0.181$ & $0.000\%$
	               	     & $0.972/0.972$ & $0.234$ & $0.000\%$ \\
		\textbf{LFR} & $0.554/0.357$ & $0.000$ & $100.0\%$ 
		             & N/A & N/A & N/A
		             & N/A & N/A & N/A \\
		\textbf{MinDiff} & $0.972/0.972$ & $0.050$ & $0.000\%$ 
		                 & N/A & N/A & N/A
		                 & N/A & N/A & N/A \\
		\textbf{DI}  & $0.972/0.972$ & $0.050$ & $0.000\%$
		             & $0.972/0.972$ & $0.181$ & $0.000\%$
	               	 & $0.972/0.972$ & $0.234$ & $0.000\%$ \\
		\textbf{Adversarial} & $0.554/0.357$ & $0.000$ & $100.0\%$ 
		                     & $0.554/0.357$ & $0.000$ & $100.0\%$
		                     & $0.554/0.357$ & $0.000$ & $100.0\%$ \\
		\textbf{FCFC} & $0.446/0.308$ & $0.000$ & $100.0\%$ 
		              & $0.446/0.308$ & $0.000$ & $100.0\%$
		              & $0.446/0.308$ & $0.000$ & $100.0\%$ \\
		\textbf{GerryFair} & $0.972/0.972$ & $0.050$ & $0.000\%$
		                   & $0.972/0.972$ & $0.181$ & $0.000\%$
	               	       & $0.972/0.972$ & $0.234$ & $0.000\%$ \\
	    \textbf{GDP} & $0.972/0.972$ & $0.050$ & $0.000\%$ 
		             & $0.972/0.972$ & $0.181$ & $0.000\%$ 
		             & $0.972/0.972$ & $0.234$ & $0.000\%$ \\
		\hline
		\textbf{\multifair} & $0.924/0.923$ & $0.038$ & $23.15\%$
		                    & $0.815/0.803$ & $0.179$ & $1.010\%$
		                    & $0.877/0.872$ & $0.231$ & $1.350\%$ \\
		\hline
		\multicolumn{10}{|c|}{\textbf{Debiasing results on {\em Adult Income} dataset}} \\
		\hline
		\multirow{3}{*}{\textbf{Method}} & \multicolumn{3}{c|}{\textbf{gender}} & \multicolumn{3}{c|}{\textbf{race}} & \multicolumn{3}{c|}{\textbf{gender \& race}} \\
		\cmidrule{2-10}
		& \textbf{Micro/Macro F1} & \textbf{Imparity} & \textbf{Reduction}
		& \textbf{Micro/Macro F1} & \textbf{Imparity} & \textbf{Reduction}
		& \textbf{Micro/Macro F1} & \textbf{Imparity} & \textbf{Reduction} \\
		\hline
		\textbf{Vanilla} & $0.830/0.762$ & $0.066$ & $0.000\%$
		                 & $0.830/0.762$ & $0.062$ & $0.000\%$
	               	     & $0.830/0.762$ & $0.083$ & $0.000\%$ \\
		\textbf{LFR} & $0.743/0.426$ & $0.000$ & $100.0\%$ 
		             & N/A & N/A & N/A
		             & N/A & N/A & N/A \\
		\textbf{MinDiff} & $0.828/0.746$ & $0.058$ & $12.06\%$ 
		                  & N/A & N/A & N/A
		                  & N/A & N/A & N/A \\
		\textbf{DI} & $0.823/0.730$ & $0.053$ & $19.85\%$ 
		            & $0.825/0.743$ & $0.056$ & $10.62\%$ 
		            & $0.823/0.736$ & $0.081$ & $2.276\%$ \\
		\textbf{Adversarial} & $0.743/0.426$ & $0.000$ & $100.0\%$ 
		                     & $0.743/0.426$ & $0.000$ & $100.0\%$ 
		                     & $0.743/0.426$ & $0.000$ & $100.0\%$  \\
		\textbf{FCFC} & $0.257/0.204$ & $0.000$ & $100.0\%$ 
		              & $0.257/0.204$ & $0.000$ & $100.0\%$ 
		              & $0.257/0.204$ & $0.000$ & $100.0\%$ \\
		\textbf{GerryFair} & $0.833/0.752$ & $0.056$ & $15.70\%$
		                   & $0.833/0.752$ & $0.067$ & $-7.664\%$
		                   & $0.797/0.710$ & $0.215$ & $-158.3\%$ \\
		\textbf{GDP} & $0.825/0.744$ & $0.055$ & $16.73\%$ 
		             & $0.827/0.749$ & $0.059$ & $6.351\%$
		             & $0.824/0.740$ & $0.075$ & $9.246\%$ \\
		\hline
		\textbf{\multifair} & $0.816/0.721$ & $0.047$ & $29.24\%$
		                    & $0.810/0.686$ & $0.042$ & $32.11\%$
		                    & $0.818/0.714$ & $0.082$ & $1.532\%$ \\
		\hline
		\multicolumn{10}{|c|}{\textbf{Debiasing results on {\em Dutch Census} dataset}} \\
		\hline
		\multirow{3}{*}{\textbf{Method}} & \multicolumn{3}{c|}{\textbf{gender}} & \multicolumn{3}{c|}{\textbf{marital status}} & \multicolumn{3}{c|}{\textbf{gender \& marital status}} \\
		\cmidrule{2-10}
		& \textbf{Micro/Macro F1} & \textbf{Imparity} & \textbf{Reduction}
		& \textbf{Micro/Macro F1} & \textbf{Imparity} & \textbf{Reduction}
		& \textbf{Micro/Macro F1} & \textbf{Imparity} & \textbf{Reduction} \\
		\hline
		\textbf{Vanilla} & $0.832/0.831$ & $0.119$ & $0.000\%$
		                 & $0.832/0.831$ & $0.079$ & $0.000\%$
	               	     & $0.832/0.831$ & $0.172$ & $0.000\%$ \\
		\textbf{LFR} & $0.521/0.342$ & $0.000$ & $100.0\%$ 
		             & N/A & N/A & N/A
		             & N/A & N/A & N/A \\
		\textbf{MinDiff} & $0.831/0.830$ & $0.107$ & $10.16\%$ 
		                  & N/A & N/A & N/A
		                  & N/A & N/A & N/A \\
		\textbf{DI} & $0.825/0.824$ & $0.104$ & $12.43\%$ 
		            & $0.830/0.830$ & $0.080$ & $-1.156\%$ 
		            & $0.814/0.811$ & $0.127$ & $26.65\%$ \\
		\textbf{Adversarial} & $0.521/0.342$ & $0.000$ & $100.0\%$
		                     & $0.521/0.342$ & $0.000$ & $100.0\%$
		                     & $0.521/0.342$ & $0.000$ & $100.0\%$ \\
		\textbf{FCFC} & $0.479/0.324$ & $0.000$ & $100.0\%$
		              & $0.479/0.324$ & $0.000$ & $100.0\%$
		              & $0.479/0.324$ & $0.000$ & $100.0\%$ \\
		\textbf{GerryFair} & $0.826/0.823$ & $0.078$ & $34.29\%$
		                   & $0.826/0.823$ & $0.070$ & $11.70\%$
		                   & $0.826/0.823$ & $0.125$ & $27.53\%$ \\
		\textbf{GDP} & $0.828/0.826$ & $0.097$ & $18.31\%$ 
		             & $0.827/0.826$ & $0.086$ & $-9.056\%$
		             & $0.827/0.825$ & $0.131$ & $23.80\%$ \\
		\hline
		\textbf{\multifair} & $0.817/0.813$ & $0.068$ & $43.08\%$
		                    & $0.815/0.811$ & $0.077$ & $2.017\%$
		                    & $0.819/0.817$ & $0.128$ & $25.65\%$ \\
		\hline
	\end{tabular}}
    \vspace{-6mm}
\end{table*}

\noindent{\bf C -- Metrics.} To answer \textbf{RQ1}, we measure the performance of classification using micro F1 and macro F1 (Micro/Macro F1). 
To answer \textbf{RQ2}, we measure to what extent the bias is reduced by the average statistical imparity (Imparity) and the relative bias reduction (Reduction) on average statistical imparity. The average statistical imparity (Imparity) is defined as $\textit{Imparity} = \textit{avg}(|\textrm{Pr}({\hat y} = c | \mathbf{x}\in g_1) - \textrm{Pr}({\hat y} = c | \mathbf{x}\in g_2)|)$ for any class label $c$ and any pair of two different demographic groups $g_1$ and $g_2$. The relative bias reduction measures the relative decrease of the imparity of the debiased outcomes $\textit{Imparity}_{\textit{debiased}}$ to the imparity of vanilla outcomes (i.e., outcomes without fairness consideration) $\textit{Imparity}_{\textit{vanilla}}$. It is computed mathematically as $\textit{Reduction} = 1 - \frac{\textit{Imparity}_{\textit{debiased}}}{\textit{Imparity}_{\textit{vanilla}}}$. Note that relative bias reduction defined above can be negative if the debiased learning outcome contains more biases than the vanilla learning outcome.
More experimental settings regarding reproducibility are provided in Appendix.

\vspace{-2mm}
\subsection{Main Results}\label{subsec:main_result}
\vspace{-1mm}
We test our proposed framework, as well as baseline methods, in three different settings: debiasing binary sensitive attribute (i.e., gender for all three datasets), debiasing non-binary sensitive attribute (i.e., race for {\em COMPAS} and {\em Adult Income}, marital status for {\em Dutch Census}) and debiasing multiple sensitive attributes (i.e., gender \& race for {\em COMPAS} and {\em Adult Income}, gender \& marital status for {\em Dutch Census}). For each dataset and each setting, we train each model on training set, then select the trained model with best bias mitigation performance on validation set and report its performance on the test set. For the vanilla model (without any fairness consideration), we report the model with the highest micro and macro F1 scores. This is because the algorithm administrators are often more concerned with maximizing the utility of classification algorithms. The results of \textit{LFR} and \textit{MinDiff} in debiasing non-binary sensitive attribute and multiple sensitive attributes are absent since they only handle binary sensitive attribute by design.

\noindent \textbf{A -- Effectiveness results.} The effectiveness results of \multifair\ and baseline methods on {\em COMPAS}, {\em Adult Income} and {\em Dutch Census} datasets are shown in Table~\ref{tab:effectiveness}. We provide additional results on visualizing the trade-off between micro F1 score and average statistical imparity in Appendix. From the tables, we have the following observations. First, Our method is the only method that can mitigate bias (i.e., Imparity and Reduction) effectively and consistently with a small degree of sacrifice to the vanilla classification performance (i.e., Micro/Macro F1) for all datasets and all settings. Second, though {\em LFR}, {\em Adversarial Debiasing} and {\em FCFC} achieves the perfect bias reduction, their classification performance is severely reduced by predicting all data samples with the same label (i.e., negative sample for {\em LFR} and {\em Adversarial Debiasing}, positive sample for {\em FCFC}). Though, in a few settings, {\em DI}, {\em GerryFair} and {\em GDP} mitigate more bias than \multifair, they either {\em amplify} the bias or fail to outperform \multifair\ in the other settings. All in all, \multifair\ achieves the best balance in reducing the bias and maintaining the classification accuracy in most cases.

\begin{figure}
    \centering
    \vspace{-1mm}
    \includegraphics[width=.45\textwidth]{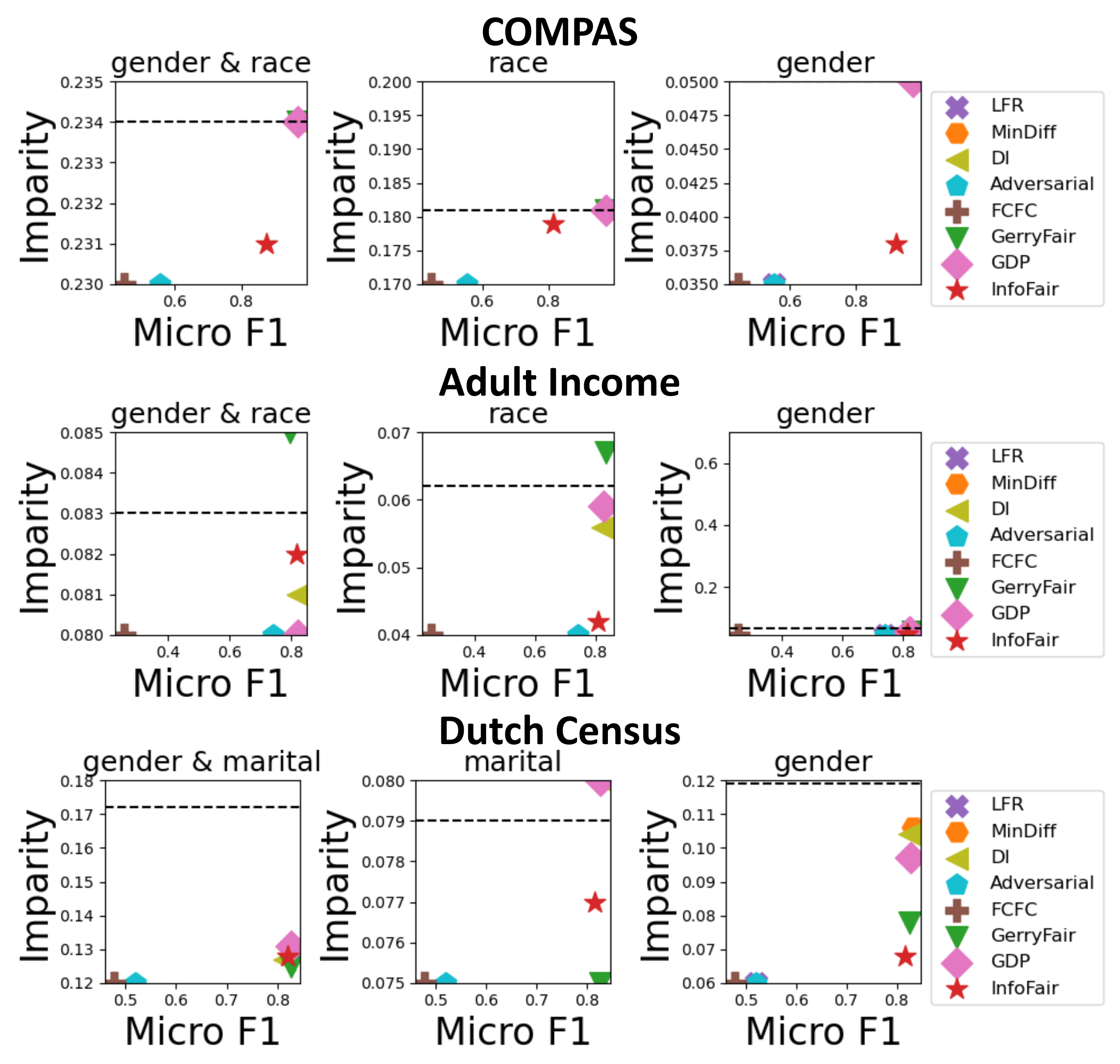}
    \vspace{-2mm}
    \caption{Trade-off between micro F1 score and average statistical imparity. Best viewed in color. Red star represents \multifair. The closer to bottom right, the better trade-off between micro F1 score and average statistical imparity. Bias is amplified by a method if its corresponding point is above the dashed line (which denotes the imparity of Vanilla).}
    \vspace{-8mm}
    \label{fig:micro_imparity}
\end{figure}

\noindent \textbf{B -- Trade-off between micro F1 and average statistical imparity.} Figure~\ref{fig:micro_imparity} shows the results of trade-off between micro F1 (Micro F1) and average statistical imparity (Imparity). From the figure, we can observe that 
\multifair\ achieves the best trade-off between accuracy and fairness (i.e., being closer to the bottom right corner in Figure~\ref{fig:micro_imparity}) in most cases.

\begin{figure}
    \centering
    \begin{subfigure}[b]{.48\textwidth}
        \centering
         \includegraphics[width=\textwidth]{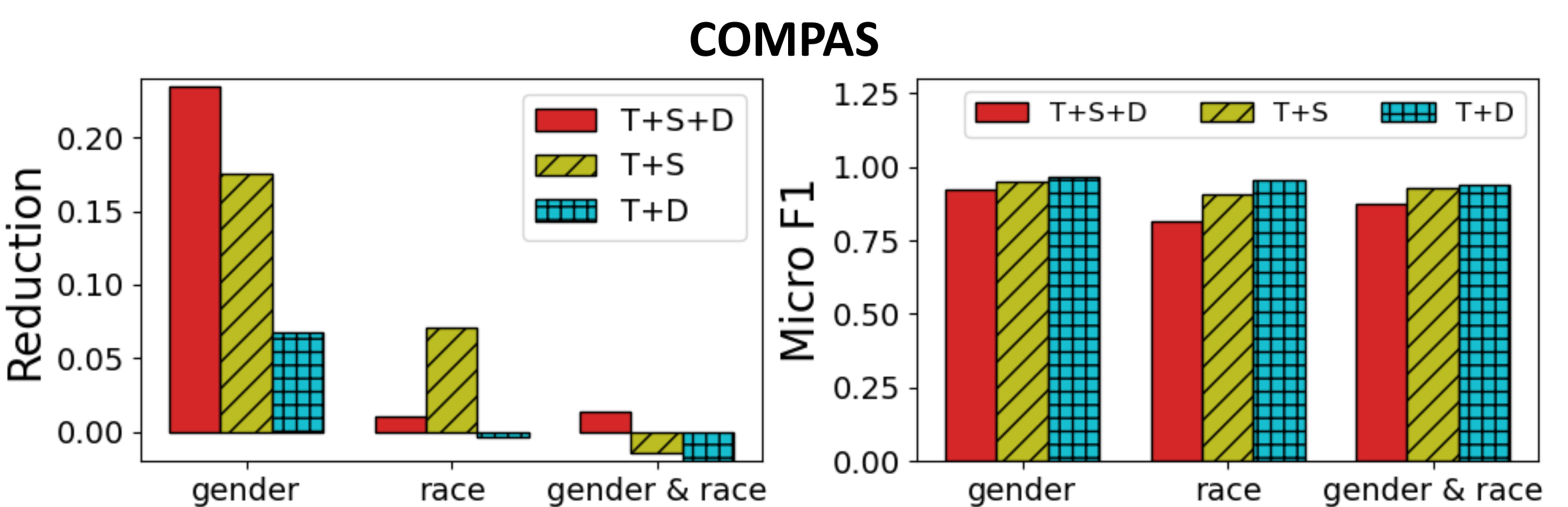}
         \vspace{-4mm}
         \label{fig:compas_micro_imparity}
    \end{subfigure}
    \newline
    \begin{subfigure}[b]{.48\textwidth}
        \centering
         \includegraphics[width=\textwidth]{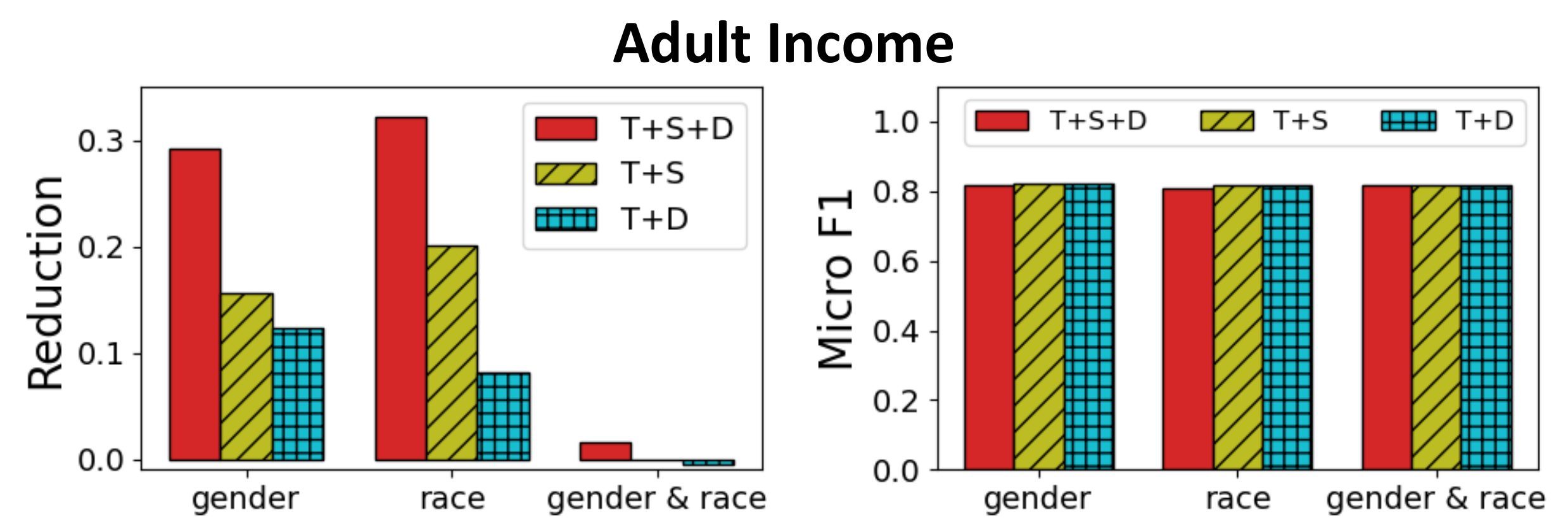}
         \vspace{-4mm}
         \label{fig:adult_micro_imparity}
    \end{subfigure}
    \newline
    \begin{subfigure}[b]{.48\textwidth}
        \centering
         \includegraphics[width=\textwidth]{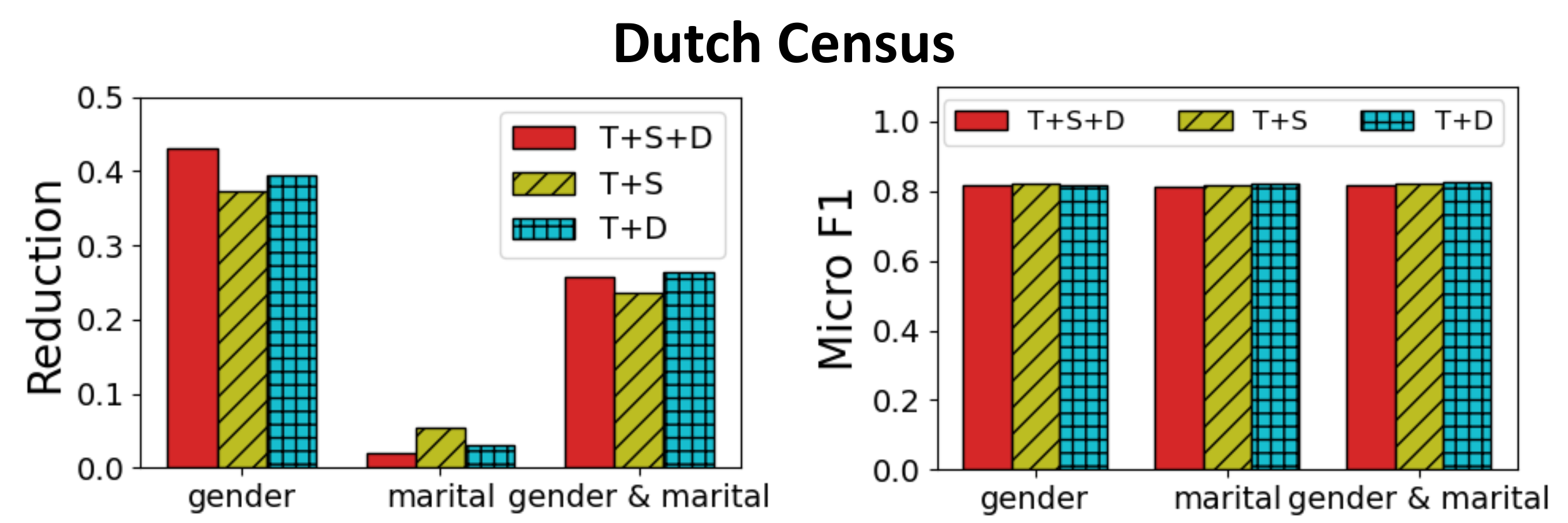}
         \vspace{-5mm}
         \label{fig:dutch_micro_imparity}
    \end{subfigure}
    \vspace{-2mm}
    \caption{Results of the ablation study on variants of objective function. Best viewed in color. Higher is better.}
    \vspace{-6mm}
    \label{fig:ablation}
\end{figure}

\vspace{-2mm}
\subsection{Ablation Study}
\vspace{-1mm}
Let $T=\mathbb{E}\big[l(\mathbf{x}; \mathbf{s}; y; \mathbf{\tilde y}; \theta)\big]$ be the empirical loss of target predictor, $S = \alpha \mathbb{E}\big[\log q_{\mathbf{s}|\mathbf{\tilde y}}\big]$ be the empirical loss of sensitive feature predictor and $D = \alpha \mathbb{E}\big[\mathbf{w}_1^T\mathbf{\tilde y} + \mathbf{w}_2^T\mathbf{s}\big]$ be the empirical loss of density ratio estimator, objective function of \multifair\ (Eq.~\eqref{eq:obj_variational}) can be written as $J=T+S+D$. To evaluate the effectiveness on optimizing the proposed variational representation of mutual information, we compare with two variants of objective function, i.e., $T+S$ and $T+D$, on the same datasets and the same set of sensitive attributes as in Section~\ref{subsec:main_result}. 
Experimental protocols and parameter settings are kept the same among all compared objective functions (i.e., $T+S+D$, $T+S$ and $T+D$).
The results of the ablation study are shown in Figure~\ref{fig:ablation}. From the figure, we observe that our objective function (i.e., $T+S+D$) can mitigate more bias than the other two variants (i.e., $T+S$ and $T+D$) in most cases. This implies that our proposed variational representation can better model the dependence between the learning outcomes and the vectorized sensitive features. 

When debiasing race on COMPAS dataset and debiasing marital status on Dutch Census dataset, we observe that the cardinalities of the demographic groups are more imbalanced. Recall the goal of sensitive feature predictor is to reduce the accuracy of predicting sensitive feature using the extracted embedding. When the demographic groups are more imbalanced, it tends to learn an embedding that contains information about a wrong demographic group to reduce its accuracy. Thus, though $T+S$ may achieve lower imparity, the statistical dependence between the extracted embeddings and the sensitive feature may not be reduced, meaning that the extracted embeddings are merely shifted to correlate with wrong demographic groups. However, with the addition of density ratio estimator, we ensure that (1) not only the sensitive feature predictor makes wrong prediction (2) but also the distribution of sensitive attribute and the extracted embeddings modeled by $S$ ($p_{\mathbf{\tilde y}} q_{\mathbf{s} | \mathbf{\tilde y}}$) are similar to its corresponding original distribution (i.e., $p_{\mathbf{\tilde y}, \mathbf{s}}$).

\vspace{-2mm}
\section{Related Work}\label{sec:rel}
\vspace{-1mm}
\noindent \textbf{A -- Group fairness} 
aims to ensure statistical-based fairness notions across the entire populations. 
It has been extensively studied in many application domains, including credit scoring~\cite{feldman2015certifying}, recidivism~\cite{dressel2018accuracy}, healthcare~\cite{zemel2013learning}, recommender systems~\cite{yao2017beyond} and natural language processing~\cite{zhao2017men}. 
Kamishima et al.~\cite{kamishima2012fairness} estimate mutual information between the learning outcome and sensitive attribute by marginalizing the output of a probabilistic discriminative model.
Zemel et al.~\cite{zemel2013learning} use a regularized approach to learn fair embeddings 
for group fairness and individual fairness.
Feldman et al.~\cite{feldman2015certifying} debias input data distribution by 
linear interpolation of original data distribution and fair data distribution.
Zhang et al.~\cite{zhang2018mitigating} propose an adversarial debiasing framework. 
Bose et al.~\cite{bose2019compositional} learn fair node representations by adversarial learning. 
However, their proposed framework could only debias multiple distinct sensitive attributes instead of multiple sensitive attributes simultaneously.
Kearns et al.~\cite{kearns2018preventing} further consider subgroup fairness from game-theoretic perspective. 
Different from \cite{kearns2018preventing}, \multifair\ directly optimizes statistical parity through mutual information minimization instead of optimizing the self-defined surrogate `fairness violation' functions using game-theoretic method. 
Zafar et al.~\cite{zafar2017fairness} ensure statistical parity by minimizing the covariance between the sensitive attribute of each data sample and its distance 
to the decision boundary of a convex margin-based classifier.
Adeli et al.~\cite{adeli2019representation} remove statistical dependence by minimizing Pearson's correlation for a convolutional neural network. 
Nevertheless, \cite{zafar2017fairness, adeli2019representation} only remove the linear dependence whereas our proposed \multifair\ removes both linear and nonlinear dependence directly.
In addition to statistical parity and disparate impact, 
Hardt et al.~\cite{hardt2016equality} propose another widely-used fairness notion named equal opportunity.
Prost et al.~\cite{prost2019toward} achieve equal false positive rate through maximum mean discrepancy (MMD) minimization. However, it can only debias with respect to binary sensitive attribute by design.
Jiang et al.~\cite{jiang2021generalized} propose generalized demographic parity for tractable calculation demographic parity with respect to continuous-valued sensitive attributes.
In terms of intersectional fairness, 
Kim et al.~\cite{kim2019multiaccuracy} propose Multiaccuracy Boost to ensure low classification error for each intersectional demographic group.
Foulds et al.~\cite{foulds2020intersectional} propose $\epsilon$-differential fairness, which ensures pairwise equal acceptance rate. 
They further 
estimate $\epsilon$-differential fairness and its corresponding uncertainty
~\cite{foulds2020bayesian}.
Morina et al.~\cite{morina2019auditing} develop the equivalence between minimizing $\epsilon$-differential fairness and minimizing a linear combination of false positive rate and false negative rate in a binary classification problem.
Ramos et al.~\cite{ramos2021reputation} ensure intersectional fairness in reputation-based ranking systems by minimizing the difference among the average reputations of a user from different demographic groups. 
Different from \cite{ramos2021reputation}, our proposed \multifair\ ensures intersectional fairness from information-theoretic perspective, and is applicable to various learning tasks as shown in Section~\ref{sec:generalization}.

\noindent \textbf{B -- Mutual information estimation} for high-dimensional data has been made possible in recent decades by analyzing variational bounds of mutual information with machine learning techniques. 
Regarding variational upper bound of mutual information, Kingma et al.~\cite{kingma2013auto} and Rezende et al.~\cite{rezende2014stochastic} almost concurrently propose Variational Auto-Encoders (VAEs) which optimizes a variational upper bound of mutual information conceptually. 
Variational lower bounds of mutual information have been extensively studied recently. Barber et al.~\cite{barber2003algorithm} propose a variational lower bound of mutual information and maximize the mutual information through moment matching. 
Belghazi et al. \cite{belghazi2018mutual} propose Mutual Information Neural Estimation (MINE) 
which maximizes Donsker-Varadhan representation of Kullback-Leibler (KL) divergence~\cite{donsker1983asymptotic}. 
In \cite{belghazi2018mutual}, MINE-$f$, a variant of MINE, is proposed to maximize the variational estimation of $f$-divergence introduced by Nguyen et al.~\cite{nguyen2010estimating}. The
same variational representation of $f$-divergence has been applied to other generative models like $f$-GAN~\cite{nowozin2016f}.
Mukherjee et al.~\cite{mukherjee2020ccmi} propose a classifier-based neural estimator for conditional mutual information named CCMI.
In addition, van den Oord et al.~\cite{oord2018representation} propose infoNCE based on noise contrastive estimation (NCE)~\cite{gutmann2010noise}. Hjelm et al.~\cite{hjelm2018learning} propose Deep Infomax (DIM) to maximize the mutual information between global representation and local regions of the input, which is further generalized to graphs~\cite{velivckovic2018deep}. 

\vspace{-2mm}
\section{Conclusion}\label{sec:con}
\vspace{-1mm}
In this paper, we study information-theoretic intersectional fairness, where we aim to simultaneously debias the learning results with respect to multiple sensitive attributes. We formally define the information-theoretic intersectional fairness 
problem by measuring the dependence between the learning results and multiple sensitive attributes as the mutual information between learning results and a joint attribute formed by these sensitive attributes. Based on that, we formulate it as an optimization problem and further propose a generic end-to-end framework, which effectively minimizes mutual information between the learning results and the joint attribute through its variational representation. We perform fair classification on three real-world datasets with the consideration of categorical sensitive attributes. The empirical evaluation results demonstrate that our proposed framework can effectively debias the classification results with respect to one or more sensitive attribute(s) with little sacrifice to the classification accuracy. Our framework is generalizable to different settings beyond the scope of fair classification with categorical sensitive attributes in our experimental evaluation. In the future, we will investigate our framework in other learning tasks (e.g., recommendation) and its effectiveness in mitigating bias for continuous-valued sensitive attributes (e.g., age, income).

\vspace{-2mm}
\section*{Acknowledgement}
This work is supported by NSF (1947135, 
2134079 
and 2147375), 
the NSF Program on Fairness in AI in collaboration with Amazon (1939725), 
DARPA (HR001121C0165), 
NIFA (2020-67021-32799), 
ARO (W911NF2110088), 
and DHS (2017-ST-061-QA0001 and 17STQAC00001-03-03). 
The content of the information in this document does not necessarily reflect the position or the policy of the Government or Amazon, and no official endorsement should be inferred.  The U.S. Government is authorized to reproduce and distribute reprints for Government purposes notwithstanding any copyright notation here on.

\vspace{-2mm}
\bibliographystyle{IEEEtran}
\bibliography{ref}

\begin{thebibliography}{10}
\providecommand{\url}[1]{#1}
\csname url@samestyle\endcsname
\providecommand{\newblock}{\relax}
\providecommand{\bibinfo}[2]{#2}
\providecommand{\BIBentrySTDinterwordspacing}{\spaceskip=0pt\relax}
\providecommand{\BIBentryALTinterwordstretchfactor}{4}
\providecommand{\BIBentryALTinterwordspacing}{\spaceskip=\fontdimen2\font plus
\BIBentryALTinterwordstretchfactor\fontdimen3\font minus
  \fontdimen4\font\relax}
\providecommand{\BIBforeignlanguage}[2]{{%
\expandafter\ifx\csname l@#1\endcsname\relax
\typeout{** WARNING: IEEEtran.bst: No hyphenation pattern has been}%
\typeout{** loaded for the language `#1'. Using the pattern for}%
\typeout{** the default language instead.}%
\else
\language=\csname l@#1\endcsname
\fi
#2}}
\providecommand{\BIBdecl}{\relax}
\BIBdecl

\bibitem{luo2017deep}
C.~Luo, D.~Wu, and D.~Wu, ``A deep learning approach for credit scoring using
  credit default swaps,'' \emph{Engineering Applications of Artificial
  Intelligence}, 2017.

\bibitem{berk2018fairness}
R.~Berk, H.~Heidari, S.~Jabbari, M.~Kearns, and A.~Roth, ``Fairness in criminal
  justice risk assessments: The state of the art,'' \emph{arXiv preprint
  arXiv:1703.09207}, 2017.

\bibitem{ahmad2018interpretable}
M.~A. Ahmad, C.~Eckert, and A.~Teredesai, ``Interpretable machine learning in
  healthcare,'' in \emph{BCB}, 2018.

\bibitem{dwork2012fairness}
C.~Dwork, M.~Hardt, T.~Pitassi, O.~Reingold, and R.~Zemel, ``Fairness through
  awareness,'' in \emph{ITCS}, 2012.

\bibitem{feldman2015certifying}
M.~Feldman, S.~A. Friedler, J.~Moeller, C.~Scheidegger, and
  S.~Venkatasubramanian, ``Certifying and removing disparate impact,'' in
  \emph{KDD}, 2015.

\bibitem{zafar2017fairness}
M.~B. Zafar, I.~Valera, M.~G. Rogriguez, and K.~P. Gummadi, ``Fairness
  constraints: Mechanisms for fair classification,'' in \emph{AISTATS}, 2017.

\bibitem{morris2000significance}
S.~B. Morris and R.~E. Lobsenz, ``Significance tests and confidence intervals
  for the adverse impact ratio,'' \emph{Personnel Psychology}, 2000.

\bibitem{hardt2016equality}
M.~Hardt, E.~Price, and N.~Srebro, ``Equality of opportunity in supervised
  learning,'' in \emph{NIPS}, 2016.

\bibitem{kearns2018preventing}
M.~Kearns, S.~Neel, A.~Roth, and Z.~S. Wu, ``Preventing fairness
  gerrymandering: Auditing and learning for subgroup fairness,'' in
  \emph{ICML}, 2018.

\bibitem{bose2019compositional}
A.~Bose and W.~Hamilton, ``Compositional fairness constraints for graph
  embeddings,'' in \emph{ICML}, 2019.

\bibitem{shannon1948mathematical}
C.~E. Shannon, ``A mathematical theory of communication,'' \emph{The Bell
  System Technical Journal}, 1948.

\bibitem{zemel2013learning}
R.~Zemel, Y.~Wu, K.~Swersky, T.~Pitassi, and C.~Dwork, ``Learning fair
  representations,'' in \emph{ICML}, 2013.

\bibitem{zhang2018mitigating}
B.~H. Zhang, B.~Lemoine, and M.~Mitchell, ``Mitigating unwanted biases with
  adversarial learning,'' in \emph{AIES}, 2018.

\bibitem{ghassami2018fairness}
A.~Ghassami, S.~Khodadadian, and N.~Kiyavash, ``Fairness in supervised
  learning: An information theoretic approach,'' in \emph{ISIT}, 2018.

\bibitem{belghazi2018mutual}
M.~I. Belghazi, A.~Baratin, S.~Rajeshwar, S.~Ozair, Y.~Bengio, A.~Courville,
  and D.~Hjelm, ``Mutual information neural estimation,'' in \emph{ICML}, 2018.

\bibitem{hjelm2018learning}
R.~D. Hjelm, A.~Fedorov, S.~Lavoie-Marchildon, K.~Grewal, P.~Bachman,
  A.~Trischler, and Y.~Bengio, ``Learning deep representations by mutual
  information estimation and maximization,'' in \emph{ICLR}, 2018.

\bibitem{mukherjee2020ccmi}
S.~Mukherjee, H.~Asnani, and S.~Kannan, ``Ccmi: Classifier based conditional
  mutual information estimation,'' in \emph{UAI}, 2020.

\bibitem{bickel2009discriminative}
S.~Bickel, M.~Br{\"u}ckner, and T.~Scheffer, ``Discriminative learning under
  covariate shift.'' \emph{Journal of Machine Learning Research}, 2009.

\bibitem{jang2017categorical}
E.~Jang, S.~Gu, and B.~Poole, ``Categorical reparameterization with
  gumbel-softmax,'' in \emph{ICLR}, 2017.

\bibitem{tishby2000information}
N.~Tishby, F.~C. Pereira, and W.~Bialek, ``The information bottleneck method,''
  \emph{arXiv preprint physics/0004057}, 2000.

\bibitem{kingma2013auto}
D.~P. Kingma and M.~Welling, ``Auto-encoding variational bayes,'' in
  \emph{ICLR}, 2014.

\bibitem{kipf2017semi}
T.~N. Kipf and M.~Welling, ``Semi-supervised classification with graph
  convolutional networks,'' in \emph{ICLR}, 2017.

\bibitem{prost2019toward}
F.~Prost, H.~Qian, Q.~Chen, E.~H. Chi, J.~Chen, and A.~Beutel, ``Toward a
  better trade-off between performance and fairness with kernel-based
  distribution matching,'' \emph{arXiv preprint arXiv:1910.11779}, 2019.

\bibitem{jiang2021generalized}
Z.~Jiang, X.~Han, C.~Fan, F.~Yang, A.~Mostafavi, and X.~Hu, ``Generalized
  demographic parity for group fairness,'' in \emph{ICLR}, 2022.

\bibitem{dressel2018accuracy}
J.~Dressel and H.~Farid, ``The accuracy, fairness, and limits of predicting
  recidivism,'' \emph{Science Advances}, 2018.

\bibitem{yao2017beyond}
S.~Yao and B.~Huang, ``Beyond parity: Fairness objectives for collaborative
  filtering,'' in \emph{NIPS}, 2017.

\bibitem{zhao2017men}
J.~Zhao, T.~Wang, M.~Yatskar, V.~Ordonez, and K.-W. Chang, ``Men also like
  shopping: Reducing gender bias amplification using corpus-level
  constraints,'' in \emph{EMNLP}, 2017.

\bibitem{kamishima2012fairness}
T.~Kamishima, S.~Akaho, H.~Asoh, and J.~Sakuma, ``Fairness-aware classifier
  with prejudice remover regularizer,'' in \emph{ECML/PKDD}, 2012, pp. 35--50.

\bibitem{adeli2019representation}
E.~Adeli, Q.~Zhao, A.~Pfefferbaum, E.~V. Sullivan, L.~Fei-Fei, J.~C. Niebles,
  and K.~M. Pohl, ``Representation learning with statistical independence to
  mitigate bias,'' \emph{arXiv preprint arXiv:1910.03676}, 2019.

\bibitem{kim2019multiaccuracy}
M.~P. Kim, A.~Ghorbani, and J.~Zou, ``Multiaccuracy: Black-box post-processing
  for fairness in classification,'' in \emph{AIES}, 2019, pp. 247--254.

\bibitem{foulds2020intersectional}
J.~R. Foulds, R.~Islam, K.~N. Keya, and S.~Pan, ``An intersectional definition
  of fairness,'' in \emph{ICDE}.\hskip 1em plus 0.5em minus 0.4em\relax IEEE,
  2020, pp. 1918--1921.

\bibitem{foulds2020bayesian}
------, ``Bayesian modeling of intersectional fairness: The variance of bias,''
  in \emph{SDM}.\hskip 1em plus 0.5em minus 0.4em\relax SIAM, 2020, pp.
  424--432.

\bibitem{morina2019auditing}
G.~Morina, V.~Oliinyk, J.~Waton, I.~Marusic, and K.~Georgatzis, ``Auditing and
  achieving intersectional fairness in classification problems,'' \emph{arXiv
  preprint arXiv:1911.01468}, 2019.

\bibitem{ramos2021reputation}
G.~Ramos, L.~Boratto, and M.~Marras, ``Reputation equity in ranking systems,''
  in \emph{CIKM}, 2021, pp. 3378--3382.

\bibitem{rezende2014stochastic}
D.~J. Rezende, S.~Mohamed, and D.~Wierstra, ``Stochastic backpropagation and
  approximate inference in deep generative models,'' in \emph{ICML}, 2014.

\bibitem{barber2003algorithm}
D.~Barber and F.~V. Agakov, ``The im algorithm: A variational approach to
  information maximization,'' in \emph{NIPS}, 2003.

\bibitem{donsker1983asymptotic}
M.~D. Donsker and S.~S. Varadhan, ``Asymptotic evaluation of certain markov
  process expectations for large time. iv,'' \emph{Communications on Pure and
  Applied Mathematics}, 1983.

\bibitem{nguyen2010estimating}
X.~Nguyen, M.~J. Wainwright, and M.~I. Jordan, ``Estimating divergence
  functionals and the likelihood ratio by convex risk minimization,''
  \emph{IEEE Transactions on Information Theory}, 2010.

\bibitem{nowozin2016f}
S.~Nowozin, B.~Cseke, and R.~Tomioka, ``f-gan: Training generative neural
  samplers using variational divergence minimization,'' in \emph{NIPS}, 2016.

\bibitem{oord2018representation}
A.~v.~d. Oord, Y.~Li, and O.~Vinyals, ``Representation learning with
  contrastive predictive coding,'' \emph{arXiv preprint arXiv:1807.03748},
  2018.

\bibitem{gutmann2010noise}
M.~Gutmann and A.~Hyv{\"a}rinen, ``Noise-contrastive estimation: A new
  estimation principle for unnormalized statistical models,'' in
  \emph{AISTATS}, 2010.

\bibitem{velivckovic2018deep}
P.~Veli{\v{c}}kovi{\'c}, W.~Fedus, W.~L. Hamilton, P.~Li{\`o}, Y.~Bengio, and
  R.~D. Hjelm, ``Deep graph infomax,'' in \emph{ICLR}, 2018.

\bibitem{kingma2014adam}
D.~P. Kingma and J.~Ba, ``Adam: {A} method for stochastic optimization,'' in
  \emph{ICLR}, 2015.

\end{thebibliography}

\balance
\vspace{-2mm}
\section*{Appendix}\label{sec:reproducibility}

\vspace{-2mm}
\subsection{Descriptions of Baseline Methods}
\vspace{-1mm}
\noindent\textbf{Learning Fair Representations (LFR)}~\cite{zemel2013learning} 
learns a set of fair prototype representations. Each data sample is first mapped to a prototype, which is used to predict fair outcome. We use the implementation by IBM AIF360 and the same grid search strategy for hyperparameters in \cite{zemel2013learning}.

\noindent\textbf{MinDiff}~\cite{prost2019toward} ensures equal false positive rate by minimizing the maximum mean discrepancy (MMD) between the two demographic groups with negative samples only. We implement our own version of MinDiff with the Gaussian kernel. The hyperparameters for the Gaussian kernel is set to be consistent with \cite{prost2019toward}. For fair comparison, we set the regularization hyperparameter to $0.1$, which is the same with the corresponding setting for \multifair.

\noindent\textbf{Disparate Impact (DI)}~\cite{feldman2015certifying} ensures disparate impact by interpolating the original data distribution with an unbiased distribution. 
For fair comparison, we set the linear interpolation coefficient, which is referred to as $\lambda$ in \cite{feldman2015certifying}, such that the interpolation ratios of \cite{feldman2015certifying} and ours are the same, i.e., $\frac{1-\lambda}{\lambda} = \frac{1}{\alpha}$.

\noindent\textbf{Adversarial Debiasing (Adversarial)}~\cite{zhang2018mitigating} 
uses an adversary to predict the sensitive attribute using the prediction from a predictor. Both the predictor and the adversary can be flexibly chosen. Since its official source code is not available, we implement the model using the same machine configurations as \multifair. For fair comparison, we switch (1) the predictor to feature extractor and target predictor in our proposed framework and (2) the adversary to sensitive feature predictor in our framework. We also set the same learning rate as our framework.

\noindent\textbf{Fair Classification with Fairness Constraints (FCFC)}~\cite{zafar2017fairness} measures the statistical imparity as the covariance between the sensitive attribute of a data sample and the distance of the corresponding data sample to the decision boundary of a linear classifier. We use the official implementation of FCFC provided by Zafar et al. and adopt their released parameter settings in our experiments.

\noindent\textbf{GerryFair}~\cite{kearns2018preventing}
ensures subgroup fairness for cost-sensitive classification through fictitious play from game-theoretic perspective. Since the relationship between $\alpha$ in \multifair\ and parameters of {\em GerryFair} is unclear, we use the default parameters provided in the officially released source code

\noindent\textbf{Generalized Demographic Parity (GDP)}~\cite{jiang2021generalized} computes the weighted total variation distance on local average prediction and global average prediction. For fair comparison, we use the official implementation, set the same backbone model for feature extraction and prediction and use the same regularization hyperparameter ($0.1$) as \multifair.

\vspace{-2mm}
\subsection{Experimental Protocol and Model Architectures} 
\vspace{-1mm}
The learning task we consider is fair classification with respect to categorical sensitive attribute(s). For all datasets, we take both non-sensitive features and sensitive features as input to the feature extractor. Regarding the model architecture, for {\em Adult Income} and {\em Dutch Census} datasets, the feature extractor is a $1$-layer MLP with hidden dimension $32$; the target predictor contains one hidden layer that calculates the log likelihood of predicting class label using the embeddings output by the feature extractor; and the sensitive feature predictor is similar to the target predictor that leverages one hidden layer to calculate the log likelihood of predicting the vectorized sensitive feature using the extracted embeddings. For {\em COMPAS} dataset, we set the feature extractor to be a $2$-layer MLP with hidden dimension $32$ in each layer, while keeping all other modules to be the same as they are for {\em Adult Income} and {\em Dutch Census} datasets.


\vspace{-2mm}
\subsection{Parameter Settings and Repeatability} 
\vspace{-1mm}
For all datasets, we set the regularizatioin parameter $\alpha=0.1$. The number of epochs for training is set to $100$ with a patience of $5$ for early stopping. Weight decay is set to $0.01$. We tune the learning rate as $0.001$ for {\em DI} and $0.0001$ for {\em MinDiff}, {\em Adversarial} and our method. All learnable model parameters are optimized with Adam optimizer~\cite{kingma2014adam}. The starting temperature for Gumbel-Softmax is set to $1$ and is divided by $2$ every $50$ epochs for annealing. To reduce randomness and enhance reproducibility, we run $5$ different initializations with random seed from $0$ to $4$.

\vspace{-2mm}
\subsection{Machine Configurations}
\vspace{-1mm}
All three datasets are publicly available online. All models (i.e., \multifair\ and baseline methods) are implemented with PyTorch 1.9.0 and are trained on a Linux server with 96 Intel Xeon Gold 6240R CPUs at 2.40 GHz and 4 Nvidia Tesla V100 SXM2 GPUs with 32 GB memory. We will release the source code upon publication.

\hide{\vspace{-3mm}
\subsection{J -- Trade-off between Micro F1 Score and Average Statistical Imparity}
The results of trade-off between micro F1 score (Micro F1) and average statistical imparity (Imparity) is shown in Figure~\ref{fig:micro_imparity}. From the figure, we can observe that, compared with other baseline methods, our method achieves the best trade-off between preserving classification accuracy and reducing bias (i.e., being closer to the bottom right corner in Figure~\ref{fig:micro_imparity}) in most cases.
}

\end{document}